\documentclass[letterpaper]{article} 
\usepackage{aaai2026}  
\usepackage{times}  
\usepackage{helvet}  
\usepackage{courier}  
\usepackage[hyphens]{url}  
\usepackage{graphicx} 
\usepackage{natbib}  
\usepackage{caption} 
\usepackage{algorithm}
\usepackage{algorithmic}

\usepackage{listings}
\usepackage{makecell}  
\usepackage[table,xcdraw]{xcolor}
\usepackage{amsmath}
\usepackage{multirow} 
\usepackage{enumitem}

\usepackage{newfloat}
\usepackage{listings}
\DeclareCaptionStyle{ruled}{labelfont=normalfont,labelsep=colon,strut=off} 
\floatstyle{ruled}
\newfloat{listing}{tb}{lst}{}
\floatname{listing}{Listing}
\title{SM3Det: A Unified Model for Multi-Modal Remote Sensing Object Detection}
\author {
    Yuxuan Li,
    Xiang Li$\dagger$,
    Yunheng Li,
    Yicheng Zhang,
    Yimian Dai, \\
    Qibin Hou,
    Ming-Ming Cheng,
    Jian Yang$\dagger$
}
\affiliations {
    PCA Lab, VCIP, Computer Science, NKU\\
    yuxuan.li.17@ucl.ac.uk, \{yunhengli, zhangyc\}@mail.nankai.edu.cn, \{xiang.li.implus, yimian.dai, houqb, cmm, csjyang\}@nankai.edu.cn~~~~~~~~$\dagger$ Corresponding Authors
}

\usepackage{bibentry}

\begin{document}

\maketitle

\begin{abstract}
With the rapid advancement of remote sensing technology, high-resolution multi-modal imagery is now more widely accessible. Conventional object detection models are trained on a single dataset, often restricted to a specific imaging modality and annotation format. 
However, such an approach overlooks the valuable shared knowledge across multi-modalities and limits the model's applicability in more versatile scenarios.
This paper introduces a \textbf{new task} called Multi-Modal Datasets and Multi-Task Object Detection (M2Det) for remote sensing, designed to accurately detect horizontal or oriented objects from any sensor modality.
This task poses challenges due to 1) the trade-offs involved in managing multi-modal modelling and 2) the complexities of multi-task optimization. To address these, we establish a benchmark dataset and propose a unified model, \textbf{SM3Det} (\underline{\textbf{S}}ingle \underline{\textbf{M}}odel for \underline{\textbf{M}}ulti-Modal datasets and \underline{\textbf{M}}ulti-Task object \underline{\textbf{Det}}ection). 
SM3Det leverages a grid-level sparse MoE backbone to enable joint knowledge learning while preserving distinct feature representations for different modalities. Furthermore, we propose a novel consistency and synchronization optimization mechanism, allowing it to effectively handle varying levels of learning difficulty across modalities and tasks.
Extensive experiments demonstrate SM3Det's effectiveness and generalizability, consistently outperforming the combination of specialized models on individual datasets. 
\end{abstract}

\begin{links}
    \link{Code}{github.com/zcablii/SM3Det}
    \link{Datasets}{www.kaggle.com/datasets/greatbird/soi-det}
    \link{Extended Version}{https://arxiv.org/pdf/2412.20665}
\end{links}

\section{Introduction}
\label{sec:intro}
\begin{figure}[!t]
  \centering
  \includegraphics[width=0.95\linewidth]{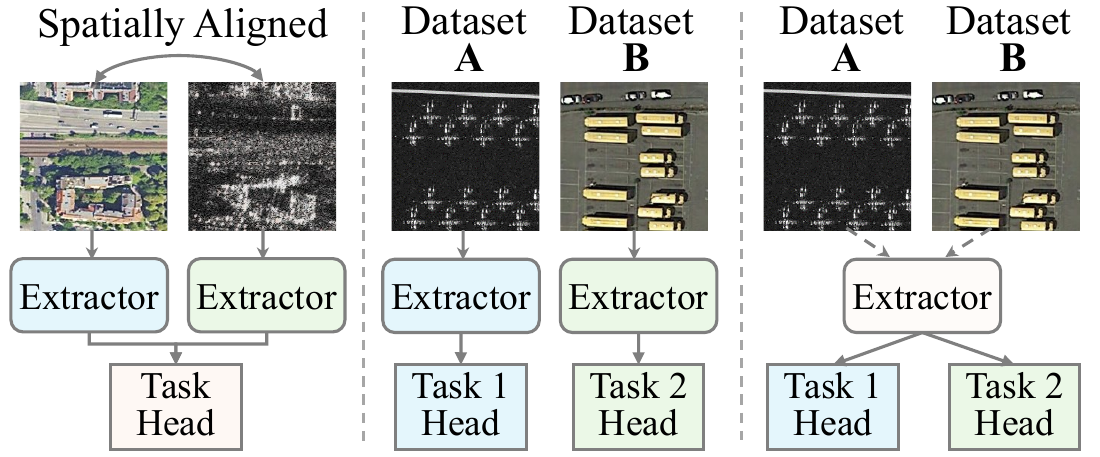}
    
  \caption{Comparison of tasks: (a) Spatially Aligned Multi-Modality, (b) Traditional Single Dataset, and (c) M2Det. M2Det aims to utilize a unified model for detecting objects in any modality, handling various detection tasks.}
  \label{fig:task_comp}
\end{figure}

Remote sensing object detection~\cite{yuan2025strip,denodetv2,Li_2025_ViTP,li2024predicting,dai2024denodet} typically involves multiple sensors employing different imaging mechanisms, resulting in diverse data modalities. Traditionally, detection models are developed for specific datasets associated with a single modality and a predefined format detection task~\cite{li2024lsknet,gwd,dai2021attentional}, as shown in Figure~\ref{fig:task_comp} (b). This conventional approach overlooks the valuable and inherent joint knowledge within a unified remote sensing context. Furthermore, airborne platforms such as UAVs and satellites often carry multiple sensors, making it critical to process images from various modalities simultaneously. Previous multi-source object detection methods~\cite{liu2021multi,zhang2024hgr,zhang2024optical} have heavily relied on scarce, impractical, and inflexible spatially well-aligned paired images and spatial alignment algorithms~\cite{devaraj2013automated,ahamed2012tower}. These methods are also limited to performing single-format detection tasks, as depicted in Figure~\ref{fig:task_comp} (a).
Thus, it is essential to develop a unified model capable of handling all modalities without requiring spatially aligned image pairs and performing multiple format detection tasks (referred to as ``multi-tasks" throughout the paper), which is not thoroughly studied. To the best of our knowledge and industrial experience, this task has the potential to serve as a \textbf{foundational technology} for emerging low-altitude economies and applications involving flying cars, drones and satellites.

To fill this research gap, we propose a new task called Multi-Modal Datasets and Multi-Task Object Detection (M2Det). M2Det aims to detect objects in any given image, regardless of its modality, and across predefined detection tasks—whether horizontal bounding boxes or oriented bounding boxes—as illustrated in Figure~\ref{fig:task_comp} (c).

The M2Det task is closely related to two key research areas: multi-dataset object detection~\cite{DA,unidet} and multi-task learning~\cite{zhang2021survey,gradnorm}. 
However, the M2Det task presents unique challenges. In traditional multi-dataset object detection, even though images may have different attributes—such as natural images and paintings—they often share similar underlying concepts (optical concepts). A simple joint training approach is effective, with a single model trained on the combined dataset typically outperforming models trained on individual datasets~\cite{DA}. 
In contrast, multi-modal datasets in remote sensing—such as RGB~\cite{dota,fair1m}, SAR~\cite{sardet100k,zhang2021sar}, IR~\cite{droneVehicle}, and multi-spectral images~\cite{Potsdam}—exhibit fundamentally different pattern concepts (as in Figure \ref{fig:experts}). While certain common knowledge may be shared across these modalities, the significant differences in data representation create a substantial modality gap, complicating the integration of information across modalities.
Additionally, remote sensing datasets often include diverse annotation types, such as horizontal~\cite{dior,sardet100k} and oriented~\cite{dota,droneVehicle} bounding boxes, further adding complexity to model learning.

These challenges may impede traditional model learning and optimization in the following ways:
\textbf{1)} Representation Constraints: A dense model that shares the same parameters across multiple tasks and modalities may encounter limitations in representation capacity, as a single set of parameters may struggle to effectively fit the diverse distributions inherent in each dataset.
\textbf{2)} Optimization Inconsistencies: The varying learning difficulties across different modalities and tasks can lead to unsynchronized optimization rates or optimization directions for various components of the model. This inconsistency can result in conflicting optimization outcomes, adversely affecting the model’s ability to achieve different loss objectives.

To address these challenges, we first establish a comprehensive benchmark dataset by merging SARDet-100K~\cite{sardet100k}, DOTA~\cite{dota}, and DroneVehicle~\cite{droneVehicle}, which collectively span SAR, optical, and infrared modalities.
Subsequently, we propose a unified model, SM3Det, tailored for the M2Det task in remote sensing, addressing the challenges from both model architecture and model optimization perspectives:

\textbf{Model Architecture:} 
We propose integrating a plug-and-play grid-level sparse Mixture of Experts (MoE) architecture into backbone networks, enabling the model to capture both shared knowledge and modality-specific representations. 
In contrast to prior multi-dataset object detection models that use hard-coded, image-level routing~\cite{DA,jain2024damex}, our approach introduces grid-level experts with dynamic routing. These experts operate on spatial grid features, allowing the model to adaptively process information at a grid level, which is crucial for object detection tasks. 

\textbf{Model Optimization:} 
We introduce a novel dynamic submodule optimization (DSO) mechanism for model optimization consistency and synchronization. 
It adaptively adjusts the learning rates of various network components based on tailored policies. 
DSO accommodates the varying learning complexities across different tasks and modalities by balancing the relative convergence rate and guaranteeing optimization direction consistency. 
Unlike traditional techniques that primarily modify loss weights or gradients—often lacking precise manipulation over specific network submodules or suffering from inefficiencies—our DSO provides fine-grained control while maintaining optimization efficiency.

Intensive experiments indicate that our unified single SM3Det model significantly outperforms individual models across all modality datasets. Our lightweight SM3Det variant not only demonstrates excellent performance but also features a substantially reduced number of parameters. Furthermore, the SM3Det model exhibits strong generalizability, enabling it to adapt to various backbones and detectors.
Our contributions are summarized as follows:
\begin{itemize}[leftmargin=1em]
  \item We introduce a new task: Multi-Modal Datasets and Multi-Task object detection in remote sensing using a unified detection model.
  \item {We propose the SM3Det model, which addresses the challenges of the M2Det task by offering innovative solutions from both model architecture and model optimization perspectives.}
  \item Extensive experiments and analyses on the established benchmark dataset demonstrate that our proposed single model is effective and outperforms individual models across all modalities.
\end{itemize}

\section{Related Work}
\label{sec:related}

\subsection{Multi-Dataset Object Detection}

Multi-dataset object detection aims to leverage a diverse collection of datasets to learn general knowledge and achieve universal object detection. 
Leveraging multiple datasets in training has proven to be a highly effective strategy for enhancing the performance of deep learning models across various applications~\cite{kapidis2021multi,zhao2020object,yan2020learning,zhang2025unichange,yang2019hierarchical}. This approach has also been widely explored in the domain of object detection. The DA network~\cite{DA}, for instance, employs specialized SE layers~\cite{SEnet} that serve as domain-specific attention mechanisms for individual datasets. Universal-RCNN~\cite{xu2020universal} introduces a partitioned detector trained across multiple datasets, integrating features through an inter-dataset graph-based attention module. Unidet~\cite{unidet} advances this concept by proposing a unified label space and underscoring the importance of batch sampling strategies.

Models trained on combined optical-concept datasets typically outperform those trained on individual datasets, as multi-dataset training can serve as a powerful form of data augmentation. However, the diverse imaging modalities in remote sensing present unique challenges for joint training. This area remains largely unexplored.

\subsection{Multi-Task Learning}
Multi-task learning involves utilizing a single model to learn multiple objectives, typically with multiple task heads and loss functions. 
In multi-task learning, various strategies~\cite{gradnorm,pareto,dtp,uncertainty} have been developed to address task imbalances and optimize learning outcomes. GradNorm~\cite{gradnorm} focuses on correcting gradient imbalances during backpropagation by adjusting the gradient sizes for each task's loss function. 
Methods like Multi-Gradient Descent Algorithm~\cite{pareto} employ Pareto optimization for gradient backpropagation, though they can be inefficient due to the additional gradient calculations required. Similar to GradNorm, DWA~\cite{DA} also uses task losses to assess convergence rates, however, it dynamically adjusts the weight of each task’s loss instead. 
Uncertainty~\cite{uncertainty} loss takes a different approach by incorporating homoscedastic uncertainty into the weighted loss function. 

Unlike loss reweighting or gradient manipulation, our method dynamically adjusts the learning rate for network submodules, enhancing multi-modal datasets and multi-task learning by maintaining optimization consistency.

\subsection{Mixture of Experts (MoE)}

MoE~\cite{moe1,moe2} leverages multiple expert networks to provide rich features. Sparse MoE\cite{sparse_MoE} further introduces sparsity, allowing for the scaling up of model size without dramatically increasing computational complexity. 
In multi-task learning, sparse MoE enables different expert networks to learn distinct discriminative features. Most sparse MoE-based multi-task methods~\cite{chen2023adamv,yang2024multi} are grounded in transformer architectures, integrating experts into vision transformer backbone blocks to selectively activate different paths during inference. 
DeepMoE~\cite{liu_improving_2020} borrow the concept of sparse MoE into CNN networks by treating the channels within each convolutional layer as experts, enhancing representational power by adaptively sparsifying and recalibrating channel features.
In multi-dataset learning, recent work~\cite{jain2024damex} employs MoE within vision transformers to route image-level features to specialized experts.

{However, sparse MoE for multi-modal datasets learning remains largely unexplored. Unlike prior methods that implement hard-coded, image-level routing~\cite{DA,jain2024damex}, 
we propose to leverage MoE into backbone networks at the feature grid level. This enables experts to effectively process spatial features, learning both shared representations and distinct patterns across modalities. }

\section{Methods}
\subsection{Task Definition}

The proposed M2Det task is designed to utilize a unified model for detecting objects in images from any modality, handling various predefined detection tasks, such as horizontal and rotated bounding boxes.
The significance of this task is evident in various real-world applications, including low-altitude economy~\cite{jiang20236g,huang2024low}, aerial surveillance~\cite{avola2021low,bozcan2020air}, earth observation~\cite{li2017earth,anderson2017earth}, and other research domains~\cite{khan2014information,jensen2016drone}. For instance, plantforms equipped with M2Det models can fully leverage available multi-modal data while benefiting from simplified version control and the seamless integration of multiple sensors without requiring model updates on the device. This significantly reduces model maintenance costs in industrial applications. Furthermore, processing images of different modalities in a single model within one mini-batch maximizes the parallel computing capabilities of GPUs, thereby enhancing computational and energy efficiency on edge devices.

\subsection{Methodological Overview}

The overall network architecture follows the classic design of multi-task learning models~\cite{DA,unidet}. It consists of a relatively heavy feature space shared component (backbone) and relatively lightweight feature space independent components (task heads). The backbone is responsible for joint representation learning, with most parameters being shared, thus ensuring parameter efficiency. The lightweight heads are separated to accommodate distinct features and task learning.
However, as discussed in Section~\ref{sec:intro}, modality and task gaps may degrade the performance of such classic multi-task models. To address this issue, we propose the SM3Det model, which consists of two parts:

\noindent \textbf{Model architecture:} {A sparse MoE backbone where experts are activated on local image features of multi-modality dataset images at the grid level.}

\noindent \textbf{Model optimization:} An efficient dynamic submodule optimization mechanism, to handle the varying learning difficulties and optimization inconsistency across multiple tasks and modalities.

\begin{figure*}[t]
  \centering
  \includegraphics[width=0.9\linewidth]{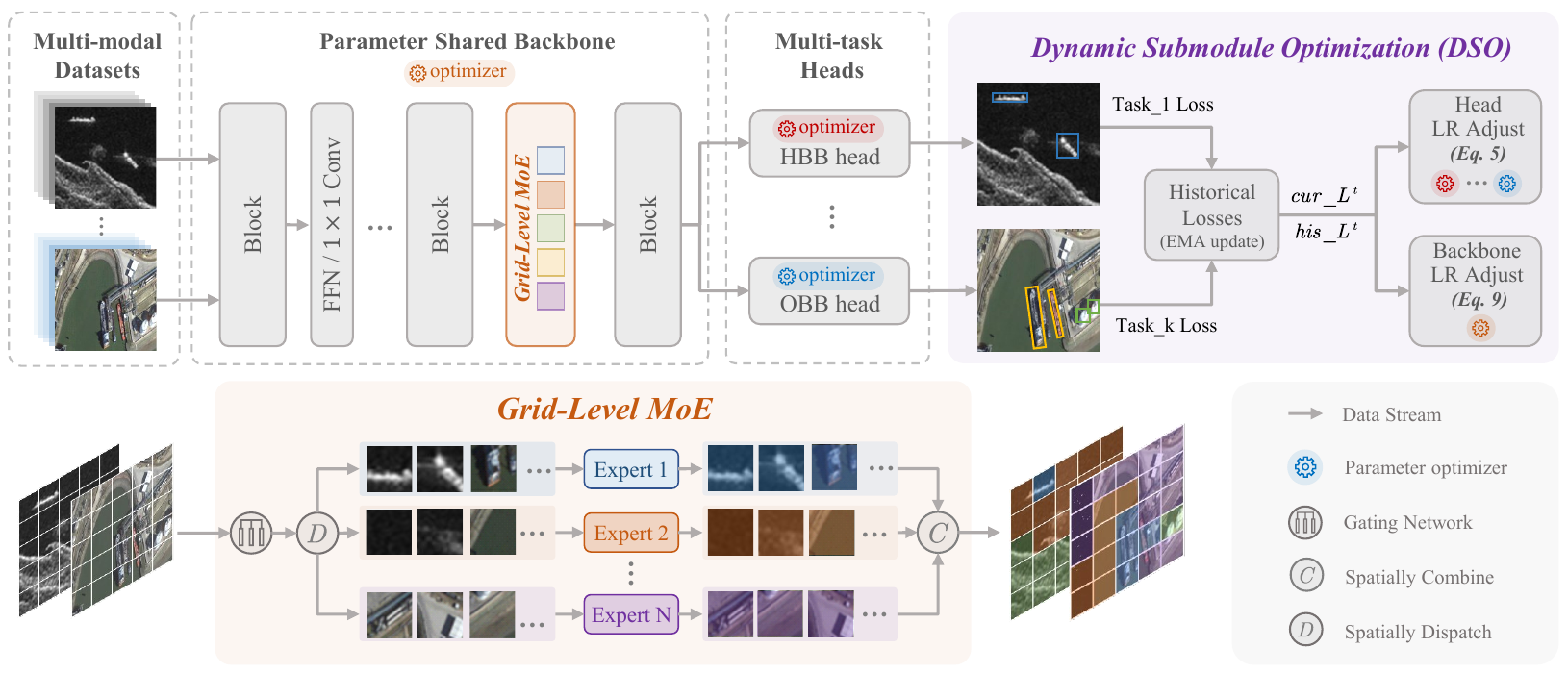}
  \caption{A conceptual illustration of SM3Det model. ``HBB'': horizontal bounding box, ``OBB'': oriented bounding box.}
  \label{fig:SM3Det}
\end{figure*}

\subsection{Grid-level MoE}

Previous approaches to multi-dataset object detection~\cite{unidet,xu2020universal} utilize dense models that leverage shared concepts among datasets to enhance joint knowledge representation. In the case of multi-modal remote sensing images, this joint knowledge also exists~\cite{sardet100k}, though it may be less explicit, with common weak cues such as shape and scale across modalities. However, due to inherent modality and task gaps, employing a dense model that utilizes the same parameters across multiple tasks and modalities can result in a congested feature/representation space, ultimately reducing the model's expressiveness. Therefore, it is essential to explore methods that leverage joint knowledge across modalities while enabling distinct representation learning for each modality to prevent feature space interference. 

Drawing inspiration from the success of Sparse MoE networks~\cite{sparse_MoE}, which are characterized by their sparsity and high capacity, we propose leveraging MoE for the M2Det task. 
For transformer-based backbones~\cite{swin,wang2022pvt}, we integrate MoE experts within the FFN components.
For modern CNNs~\cite{liu2022convnet,van,lsknet}, which often employ 1$\times$1 convolutions~\cite{lin2013network} for feature interaction or dimensionality reduction/expansion, we introduce sparse experts to enhance these layers.
{Unlike previous transformer-based detectors that route an entire image’s features through a single expert~\cite{jain2024damex}, our design allows experts to operate on local grid features within the backbone. This approach ensures that experts process similar spatial patterns across modalities, facilitating shared representation learning. Simultaneously, multiple experts capture distinct patterns across modalities, enabling independent representation learning.} Specifically, for the local spatial input feature $x_{ij}$ at the $i$-th row and $j$-th column of a deep image feature, the output feature $f_{MoE} (x_{ij})$ after the MoE layer is:
\begin{align} 
 &f_{MoE} (x_{ij}) = \sum_{n=1}^{N} G_n(x_{ij}) \cdot Conv^{1\times 1}_{n}(x_{ij}),
\label{eqn:1} \\ 
&G(x_{ij}) = \text{TOP}_{k} \left(\text{Softmax} \left(\frac{E^T W x_{ij}}{\tau \|Wx_{ij}\| \|E\|}\right)\right),
\label{eqn:2}
\end{align}
where $N$ is the total number of experts, $G$ is the gating function and $Conv^{1\times 1}_{n}$ is the $n$-th 1$\times$1 convolutional expert. Each expert has a representation embedding in the matrix $E$. The input feature $x$ is first transformed by the matrix $W$. The product of $Wx$ is then compared with each expert embedding in $E$ to calculate the similarity. This comparison is then normalized by the product of the norms of $Wx$ and $E$, ensuring scale-invariance. The similarity scores are passed through a $Softmax$ function, converting them into a probability distribution. This means the gating function assigns a probability to each expert, indicating its relevance to the input feature $x$. Finally, the $\text{TOP}_k$ operator selects the top-$k$ experts with the highest probabilities. It reweights each expert by assigning the $Softmax$ probability to the top-$k$ experts, setting the rest to zero. This step sparsifies the model by focusing only on a small subset of experts, reducing computational complexity and enhancing the model's expressiveness to handle diverse tasks and modalities.  


In summary, $f_{MoE}(x_{ij})$ is a weighted sum of the outputs from top-$k$ experts. The weights are determined by the gating function $G$, which dynamically selects the most relevant expert(s) for each local feature. The MoE creates a sparser feature space in the backbone model. By focusing on local patterns, the model can learn independently to model multiple modalities and local object patterns. Our design effectively addresses the challenges of crowded feature spaces and enhances the expressiveness of the model.

In practical implementation, to fully utilize the pretrained backbone weights, we initialize the weights of added experts by duplicating the corresponding pretrained ${1 \times 1}$ convolutional layers' weights before downstream model fine-tuning, ensuring all experts can be evenly chosen at the beginning of fine-tuning. For the task heads, we maintain simplicity and adhere to the existing design of task heads as in~\cite{DA,unidet}.

\subsection{Dynamic Submodule Optimization (DSO)}
In multi-modal, multi-dataset, and multi-task object detection tasks, one primary challenge is the varying learning difficulties~\cite{uncertainty,gradnorm} across modalities and tasks. The variation can cause unsynchronized optimization rates and inconsistent optimization directions~\cite{nakano2021cross}, leading to conflicting objectives among different loss functions. To address this problem, we propose a novel Dynamic Submodule Optimization (DSO) mechanism to manage the differing learning difficulties across tasks and modalities.

DSO takes each task head's loss as indicator to determine the current convergence rate of each task and the overall optimization direction of the network, adjusting the learning rate (LR) accordingly. Specifically, one policy is for the LR of each task head submodule (non-shared weights) to balance each task's relative convergence rate, and another policy is for the backbone submodule (shared weights) to ensure optimization direction consistency.

We denote the training loss from the iteration $i$ of task $t$ as $cur\_L^t_i$. Each task's loss maintains an exponential moving average (EMA) value as the smoothed historical statistic, denoted as $his\_L^t_i$, i.e.,
\begin{align}
{his\_L}^t_i = \alpha \cdot {cur\_L}^t_i + (1-\alpha)\cdot {his\_L}^t_{i-1} \text{ .}
\label{eqn:5}
\end{align}

\noindent\textbf{For the head submodule's LR adjustment}, we use the ratio of $his\_L$ to $cur\_L$ as the inverse of the convergence rate for iteration $i$ of task $t$ as:

\begin{align}
w^t_i = \frac{{his\_L}^t_i}{{cur\_L}^t_i}\text{ .}
\label{eqn:6}
\end{align}

The $Softmax$ with temperature $\theta$ is then used to reweight the LR of the corresponding network task head, aiming to balance the convergence speed of each task. The reweighting factor $\lambda^t_i$ for task $t$ at training iteration $i$ is denoted as:
\begin{align}
\lambda^t_i = \frac{T\cdot e^{w^t_i / \theta}}{\sum\nolimits^T_k e^{w^k_i / \theta}}\text{ ,}
\label{eqn:7}
\end{align}
where $T$ is the total number of tasks. As a result, a relatively large value of ${cur\_L}^t_i$ indicates faster convergence for task $t$, leading to a smaller $w^t_i$ and, consequently, a lower reweighting factor $\lambda^t_i$ to prevent overly rapid convergence. Conversely, a smaller value of ${cur\_L}^t_i$ results in a larger $\lambda^t_i$. This strategy ensures that the convergence rate of each task remains balanced throughout training.

\begin{figure}[!t]
  \centering
  \includegraphics[width=0.56\linewidth]{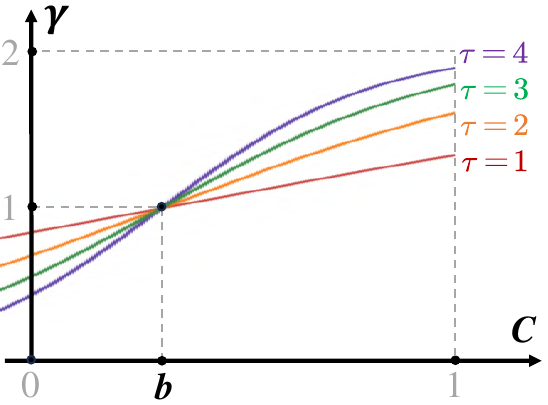}
  \caption{Reweighting curves for various temperature ($\tau$).}
  \label{fig:dla_curve}
\end{figure}

\noindent\textbf{For the backbone submodule's LR adjustment}, the reweighting is based on the historical consistency of each loss. To measure the training convergence consistency, we define a consistency score $C$ based on $cur\_L$ and $his\_L$. Specifically, $cur\_L$ and $his\_L$ are first converted into probability distributions using the function $P$, which employs a simple $Softmax$ function:

\begin{align}
P(L) &= Softmax(L).
\label{eqn:7.5}
\end{align}

Next, the Kullback-Leibler divergence, $D_{KL}$, is calculated to evaluate whether the current losses from each task remain stable and consistent with their historical values:

\begin{align}  
C &= 1 - D_{KL}\left(~P(cur\_L)~\|~P(his\_L)~\right) \\
  &= 1- \sum\nolimits^{T}_{t} P(cur\_L^t) \cdot \log \frac{P(cur\_L^t)}{P(his\_L^t)}~, ~~~~~~~~~
\label{eqn:8}
\end{align}
therefore $C$ is in the range of $(-\infty,1]$.
A larger $C$ indicates that the relative values of the current iteration losses are similar to their historical values, suggesting that the current batch of samples stabilizes the network updates. In this case, the LR has to be increased to make the network converge faster. Conversely, a lower $C$ indicates instability, suggesting that the current samples make some tasks more difficult and others easier to learn compared to the previous average state. 
If the network updates the shared weights too aggressively in such cases, the network will be optimized in the direction of the harder task of the current iteration, which might harm the easier tasks. Therefore, the network should update cautiously to reduce the LR.

To balance this, we propose dynamically reweighting the shared weight backbone with the following policy:
\begin{align}
\gamma_i &= ~2 \cdot Sigmoid((C - b) \cdot \tau) \\
         &= ~\frac{2}{~1+e^{-(C-b)\cdot \tau}~}~.
\label{eqn:9}
\end{align}

The scalar factor of 2 ensures the reweighted value after the sigmoid function is in the range of (0, 2). $b$ is the hyperparameter, bias, which can be interpreted as the reweithing threshold, i.e. when the $C$ is $b$, the reweight is 1. $\tau$ is the temperature for value sensitivity adjustment. The reweighting curves for various temperatures and the relation between $b$ and $C$ are demonstrated in Figure~\ref{fig:dla_curve}.

\section{Experiments and Analysis}

To train and evaluate models for the M2Det task, we establish a new benchmark dataset by merging three detection datasets: SARDet-100K~\cite{sardet100k}, DOTA-v1.0~\cite{dota}, and DroneVehicle~\cite{droneVehicle}, which correspond to SAR, optical, and infrared modalities, respectively. We refer to this combined dataset as the SOI-Det dataset. More detailed dataset description and implementation details can be found in the Appendix. In the main results and ablation studies, ConvNext-T is used as the default backbone unless otherwise specified.

\begin{table*}[t]
  \centering
\begin{tabular}{c|cc|cccc}
Model                                                                                  & {FLOPs}             & \#P            & Test on      & \textbf{mAP}    & @50    & @75    \\ 
\hline \Xhline{1pt}
3 models                        &      403G       &  126M     & \cellcolor[HTML]{EFEFEF}Overall      & \cellcolor[HTML]{EFEFEF}48.23 & \cellcolor[HTML]{EFEFEF}79.39 & \cellcolor[HTML]{EFEFEF}51.26 \\ \hline
GFL                 &        131G     &    36M     & {SARDet-100K}    & 57.31 & 87.44 & 61.99 \\
O-RCNN      &        136G     &    45M     & {DOTA}        & 45.31 & 77.70  & 46.45 \\
O-RCNN        &        136G     &    45M     & {DroneVehicle} & 46.09 & 74.78 & 52.79 \\ \Xhline{1pt}
\multirow{4}{*}{\begin{tabular}[c]{@{}c@{}} Simple\\      Joint\\   Training\end{tabular}} & \multirow{4}{*}{403G} & \multirow{4}{*}{66M} & \cellcolor[HTML]{EFEFEF}Overall      & \cellcolor[HTML]{EFEFEF}47.05 &  \cellcolor[HTML]{EFEFEF}77.56 &   \cellcolor[HTML]{EFEFEF}50.11 \\ \cline{4-7} 
                                                                                       &                   &                   & {SARDet-100K}    & 53.46   &84.11   &  57.29      \\
                                                                                       &                   &                   & {DOTA}         & 45.18   & 76.37   & 46.78   \\
                                                                                       &                   &                   & {DroneVehicle} & 44.99   & 73.28   & 51.50   \\ \hline
\multirow{4}{*}{\begin{tabular}[c]{@{}c@{}} DA \\ {+ConvNext-T} \end{tabular}} & \multirow{4}{*}{403G} & \multirow{4}{*}{66M} & \cellcolor[HTML]{EFEFEF}Overall      & \cellcolor[HTML]{EFEFEF}48.37      & \cellcolor[HTML]{EFEFEF}79.76      & \cellcolor[HTML]{EFEFEF}51.66     \\ \cline{4-7}
                                                                                       &                   &                   & {SARDet-100K}    &     53.86   &  84.93      &     58.09   \\
                                                                                       &                   &                   & {DOTA}         &  46.23  & {78.47}  &    47.58    \\
                                                                                       &                   &                   & {DroneVehicle} &   48.21  & 77.43   &   56.16   \\ \hline
\multirow{4}{*}{\begin{tabular}[c]{@{}c@{}}UniDet \\ {(Partitioned)}\end{tabular}}       & \multirow{4}{*}{403G} & \multirow{4}{*}{66M} & \cellcolor[HTML]{EFEFEF}Overall      & 48.47\cellcolor[HTML]{EFEFEF}       & 79.55\cellcolor[HTML]{EFEFEF}       & 52.01\cellcolor[HTML]{EFEFEF}       \\ \cline{4-7}
                                                                                       &                   &                   & {SARDet-100K}    &  53.81 &  84.70 & 57.43 \\
                                                                                       &                   &                   & {DOTA}         &  {46.49} & {78.28}  &  {48.59 }\\
                                                                                       &                   &                   & {DroneVehicle} & 47.99  & 77.17 & 55.74 \\ \hline
\multirow{4}{*}{\begin{tabular}[c]{@{}c@{}} Uncertainty \\ loss \end{tabular}} & \multirow{4}{*}{403G} & \multirow{4}{*}{66M} & \cellcolor[HTML]{EFEFEF}Overall      & \cellcolor[HTML]{EFEFEF}48.79       & \cellcolor[HTML]{EFEFEF}79.99     & \cellcolor[HTML]{EFEFEF}52.50      \\ \cline{4-7} 
                                                                                       &                   &                   & {SARDet-100K}    &   53.43  & 84.81  & 57.41   \\
                                                                                       &                   &                   & {DOTA}         &   {46.94}  &  {78.73} & {49.08}    \\
                                                                                       &                   &                   & {DroneVehicle} &   {48.78} &  {77.96}  &   {56.88}   \\ \Xhline{1pt}
\multirow{4}{*}{\begin{tabular}[c]{@{}c@{}}SM3Det \\  {(DSO only)} \end{tabular}}      & \multirow{4}{*}{403G} & \multirow{4}{*}{66M} & \cellcolor[HTML]{EFEFEF}Overall      & \cellcolor[HTML]{EFEFEF}\textcolor{blue}{49.40} & \cellcolor[HTML]{EFEFEF}\textcolor{blue}{80.19} & \cellcolor[HTML]{EFEFEF}\textcolor{blue}{52.93} \\ \cline{4-7}
                                                                                       &                   &                   & {SARDet-100K}    & {58.54}  & {88.59}  & {62.67} \\
                                                                                       &                   &                   & {DOTA}         & 46.18  & 77.86  & 47.95  \\
                                                                                       &                   &                   & {DroneVehicle} & 48.09  &  77.09 &  56.20 \\ \hline
\multirow{4}{*}{\begin{tabular}[c]{@{}c@{}}SM3Det  \end{tabular}}     & \multirow{4}{*}{487G} & \multirow{4}{*}{178M} & \cellcolor[HTML]{EFEFEF}Overall      & \cellcolor[HTML]{EFEFEF}\textcolor{red}{50.20}       & \cellcolor[HTML]{EFEFEF}\textcolor{red}{80.68}       & \cellcolor[HTML]{EFEFEF}\textcolor{red}{53.79}       \\ \cline{4-7}
                                                                                       &                   &                   & {SARDet-100K}  & {60.64}  & {89.94} & {65.06}  \\
                                                                                       &                   &                   & {DOTA}      &  46.47   &  77.88  & 48.24   \\
                                                                                       &                   &                   & {DroneVehicle} & {48.87}  & {77.99}  & {56.90}  
\end{tabular}
  \caption{Model performance comparison on the SOI-Det dataset (SARDet-100K + DOTA + DroneVehicle). The proposed SM3Det model outperforms individual models and other SOTA models. }
  \label{tab:main}
\end{table*}

\subsection{Main Results}
We evaluate the performance of our proposed SM3Det model against individual dataset training, simple joint training, and three SOTA methods that can be adapted for this task: UniDet~\cite{unidet} with a partitioned head, the DA network~\cite{DA} implemented within the ConvNext-T backbone, and uncertainty loss~\cite{uncertainty} implemented upon UniDet. The main results are presented in Table \ref{tab:main}.

It can be observed that simple joint training of the three multi-modality datasets—i.e., merely merging the datasets and using a model with a shared backbone and separate task heads, along with a random data sampling strategy—results in a significant performance drop. This phenomenon highlights the increased challenge of this task compared to multi-dataset training for general object detection, where simple joint training typically enhances the performance of individual datasets~\cite{DA,unidet}. The previous SOTA methods, {UniDet~\cite{unidet}, DA~\cite{DA} and uncertainty loss~\cite{uncertainty}}, barely exceed the baseline by a small margin. In contrast, our proposed SM3Det model significantly improves overall mAP performance from 48.23 to 50.20, an increase of 1.97 mAP. To be noticed, our lightweight version of SM3Det which only incorporates DSO but without MoE structures, also easily outperforms other SOTA methods.

\begin{table*}[!t]
  \centering 
\begin{tabular}{c|c|ccccc|ccc|cc}
MoE ($N$, k) & w/o MoE & 2, 2 & 4, 2 & 6, 2 & 8, 2 & 10, 2 & 8, 1 & 8, 2 & 8, 3 & \begin{tabular}[c]{@{}c@{}} 8, 2\\ Image-Level\end{tabular} & \begin{tabular}[c]{@{}c@{}}\hspace{-12pt} 8, 2\\ \hspace{-10pt} Grid-Level \end{tabular} \\ \Xhline{1pt}
FLOPs (G) & 403 & 469 & 469 & 469 & 469 & 469 & 403 & 469 & 531 & 487 & \hspace{-12pt}487\\
\#P (M) & 66 & 82 & 113 & 142 & 174 & 205 & 174 & 174 & 174& 178 & \hspace{-12pt}178 \\ \hline
\textbf{mAP} & 48.51 & 48.94 & 49.11 & 49.13 & \textbf{49.31} & 49.24 & 49.05 & \textbf{49.31} & 49.13 & 48.25 & \hspace{-12pt}\textbf{50.20}\\
@50 & 79.70 & 80.25 & 80.10 & 79.74 & \textbf{80.26} & 80.18 & 79.72 & \textbf{80.26} & 79.98 & 79.10& \hspace{-12pt}\textbf{80.68}\\
@75 & 51.78 & 52.01 & 52.13 & 52.76 & 52.84 & 52.79 & 52.30 & 52.84 & \textbf{52.77}& 51.31 & \hspace{-12pt}\textbf{53.79}
\end{tabular}
\caption{Experiments on the MoE backbone with varying numbers of experts and top-$k$ selection configurations. Experts are applied only to the even-indexed layers of the last two stages for validation efficiency, except for the last 2 columns. $N$: number of experts to add. k: number of experts to activate. The optimal configuration balancing performance and computational efficiency is identified as 8 experts with a top-$k$ value of 2.} 
  \label{tab:MoE_experts}
\end{table*}

\begin{figure}[!t]
  \centering
  \includegraphics[width=0.8\linewidth]{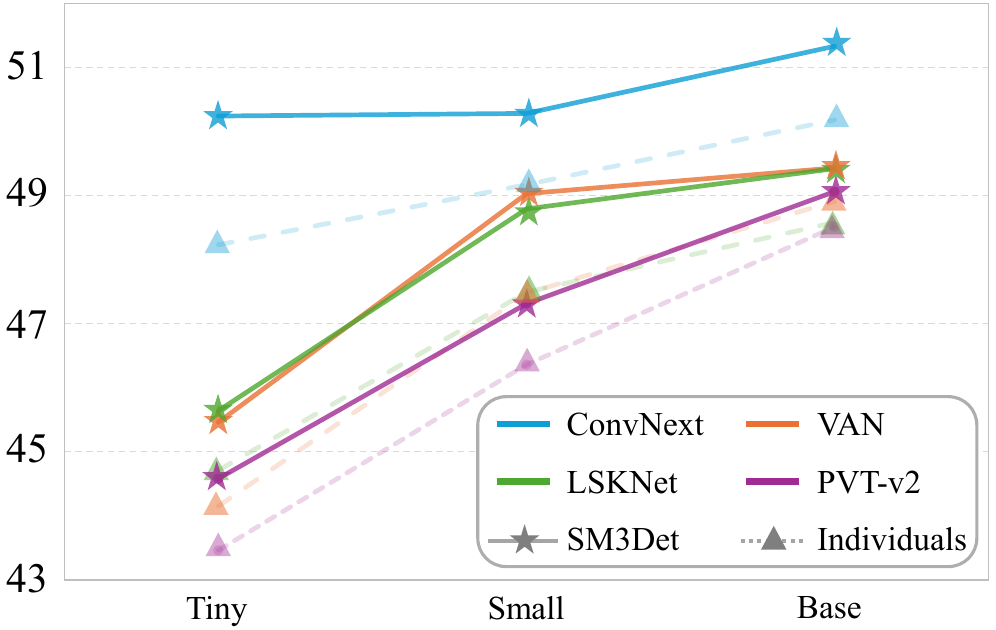}
   
  \caption{SM3Det on different backbones.}
  \label{fig:diff_backbone}
\end{figure}

\begin{figure}[!t]
  \centering
  \includegraphics[width=0.85\linewidth]{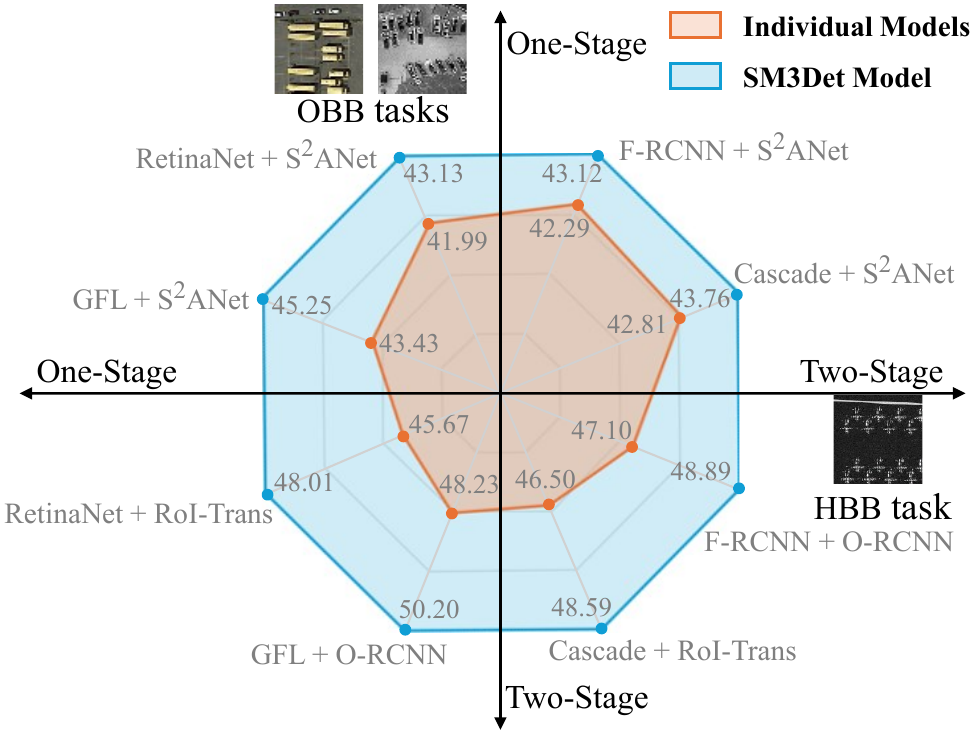}
   
  \caption{SM3Det on different detector heads. }
  \label{fig:diff_framework}
\end{figure}

To assess the generalization capability of SM3Det, we evaluate its performance across different backbones and detectors. As illustrated in Figure~\ref{fig:diff_backbone}, SM3Det significantly outperforms individual models across various modern convolutional backbones, including ConvNext~\cite{liu2022convnet}, VAN~\cite{van}, LSKNet~\cite{lsknet} and PVT-v2~\cite{wang2022pvt}. The model also exhibits reasonable scalability as the model size increases.
We also evaluate SM3Det with different detectors. Since both the optical dataset (DOTA) and the infrared dataset (DroneVehicle) involve OBB regression tasks, we use the same head network structure in our model. In contrast, for the SAR dataset (SARDet-100K), which involves an HBB regression task, we implement a standard horizontal object detection head. Figure~\ref{fig:diff_framework} shows our evaluation of SM3Det on one-stage (RetinaNet~\cite{retina}, GFL~\cite{gfl} and S$^2$ANet~\cite{s2anet}) and two-stage (F-RCNN~\cite{frcnn}, Cascade F-RCNN~\cite{cascade}, O-RCNN~\cite{orientedrcnn} and RoI-Transformer~\cite{roi_trans}) detector combinations. The results consistently demonstrate that SM3Det significantly outperforms individual models across all detector combinations.

\subsection{Ablation Study and Analysis}

\noindent\textbf{Expert Number and top-$k$ Number.}
In sparse MoE architecture, the number of experts to add ($N$) and the top-$k$ value play crucial roles in determining the model's performance and efficiency. Increasing $N$ generally enhances the model's representation capacity, while a higher top-$k$ value allows for more specialized knowledge to be applied to each input. However, these enhancements come at the cost of a larger model size, increased computational complexity, and potentially requiring more training data to ensure that each expert is adequately trained. Therefore, selecting the appropriate number of experts and top-$k$ value is critical for achieving an optimal balance between model performance and computational efficiency.
The results in Table~\ref{tab:MoE_experts} underscore the importance of tuning the number of experts and the top-$k$ value in a sparse MoE architecture. 
It reveals that the optimal configuration for this sparse MoE architecture in terms of balancing performance and computational efficiency is 8 experts with a top-2 experts. This configuration maximizes the model's ability to learn from diverse inputs without introducing unnecessary complexity or overfitting.

\noindent\textbf{Image-level v.s. Grid-level MoE.}
{In Table~\ref{tab:MoE_experts}, the grid-level MoE outperforms the image-level counterpart, indicating that grid-level experts more effectively capture spatial variations across different objects in multi-modal images. By processing features at a finer spatial granularity, experts are more attuned to object localization, making grid-level MoE particularly well-suited for object detection tasks.}

\begin{table*}[!th]

  \centering 
  
\begin{tabular}{c|cccc|ccc|ccc}
$\tau$, b  & 3, 0.3 & 3, 0.4 & 3, 0.5 & 3, 0.6 & 2, 0.4 & 3, 0.4 & 4, 0.4& {w/o DSO} & {w/o Head policy} & {w/o Backbone policy}  \\ \Xhline{1pt}
\textbf{mAP} & 50.14 & \textbf{50.20} & 50.07 & 50.03 & 49.92 & \textbf{50.20} & 50.03  & 49.47 & 49.86  & 50.11\\
@50 & 80.61 & \textbf{80.68} & 80.66 & 80.61 & 80.55 & \textbf{80.68} & 80.44  & 80.33 & 80.53  & 80.66\\
@75 & \textbf{53.81} & 53.79 & 54.00 & 53.98 & 53.56 & 53.79 & 53.79 & 52.98 & 53.44  & 53.70
\end{tabular}
  \caption{Experiments on the DSO method with varying temperature ($\tau$) and bias ($b$). DSO is not sensitive to bias $b$. } 
  \label{tab:dyn_lr_Tb}
\end{table*}

\begin{figure}[!t]
  \centering
  \includegraphics[width=\linewidth]{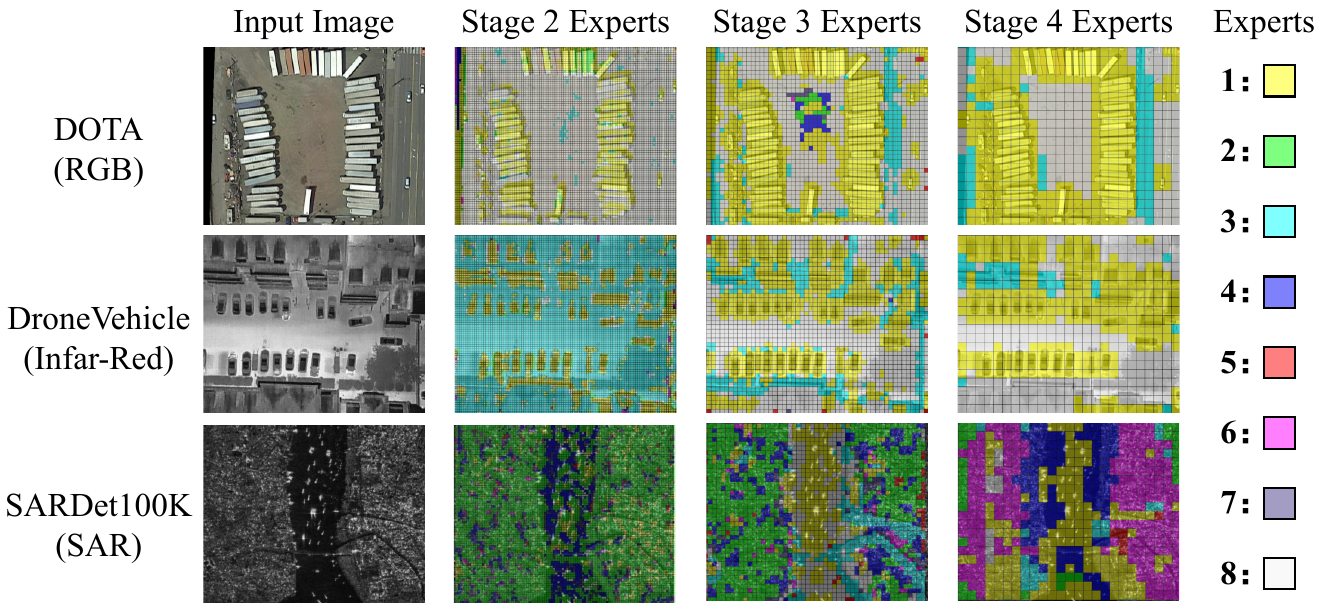}
  \caption{Visualization of grid expert activation across the last three stages of a well-tuned backbone on SAR, RGB, and IF images. Each square grid represents the receptive field at a given stage, with different colors indicating the local grid areas processed by distinct experts. The top-1 selected experts for each grid are shown. Each expert specializes in processing unique local patterns and semantics.}
  \label{fig:experts}
\end{figure}

\noindent\textbf{Grid-level Experts Activation Behaviour Analysis.}
We visualize the selection results for each grid area across the last three stages of a well-tuned ConvNext-T backbone. In this visualization, each square grid represents the corresponding receptive field of that stage, with local deep features processed by different experts indicated by distinct colours. The top-1 selected experts are illustrated in Figure~\ref{fig:experts}.
For both RGB and IR images, a consistent pattern emerges: expert 1 predominantly processes salient objects, while Expert 3 focuses on background patches across all three stages. In contrast, the situation is more complex for SAR images. Particularly at stage 4, three experts (Expert 1, Expert 4, and Expert 6) are responsible for processing background areas, with Expert 1 also handling ship objects. Additional visualizations and detailed MoE behaviour analysis can be found in the \textbf{Supplementary Material}.

\noindent\textbf{DSO hyperparamers.} 
We conduct an ablation study on each component of the proposed DSO method, as well as the sensitivity of its two key hyperparameters. The results are summarized in Table~\ref{tab:dyn_lr_Tb}. Omitting the learning rate adjustment for either the head or backbone leads to significant performance degradation. The bias parameter $b$ and temperature $\tau$ dynamically adjust learning rates to account for varying task and modality difficulties. Specifically, $b$ serves as a reweighting balance point, meaning when the calculated consistency score equals $b$, the reweighting factor is 1. A bias value of $b = 0.4$ proved optimal when the temperature was fixed at 3, striking a good balance in learning rate adjustments. Notably, variations in $b$ did not significantly impact performance, indicating that the method is robust to changes in bias.
Regarding temperature,$\tau$, influences the reweighting curve in both the network head and the backbone's learning rate adjustment mechanism. Larger values result in sharper, more sensitive adjustments. A temperature of $\tau = 3$ provided the best balance between stability and responsiveness.
In summary, the $\tau = 3$ and $b = 0.4$ yielded the best performance, effectively managing learning rate adjustments across diverse tasks and datasets.

\section{Limitation and Future Work}
An important modality in remote sensing is multi-spectrum imaging. However, due to the limited availability of large-scale multi-spectrum object detection datasets, we could not include such datasets in our experiments. To the best of our knowledge, the proposed M2Det task holds significant potential as a {foundational technology} for emerging low-altitude economies (e.g., flying cars and drones). 
Moreover, our model designs and observations can extend beyond remote sensing to other scenarios involving multiple modalities or the joint training of diverse datasets, including medical imaging~\cite{elangovan2016medical} (X-Ray, NMR and CT) and autonomous driving~\cite{feng2020deep} (camera, LiDAR and Radar). These applications present exciting avenues for future research.

\section{Conclusion}
In conclusion, this paper introduces a new and challenging task of Multi-Modal Datasets and Multi-Task Object Detection in remote sensing. To tackle this, we developed the SM3Det model, integrating a novel grid-level MoE approach and a dynamic submodule optimization mechanism. Intensive experiments and thorough analysis demonstrate SM3Det's strong performance and generalizability.

\newpage
\bibliography{aaai2026}

@InProceedings{lsknet,
    author    = {Li, Yuxuan and Hou, Qibin and Zheng, Zhaohui and Cheng, Ming-Ming and Yang, Jian and Li, Xiang},
    title     = {Large Selective Kernel Network for Remote Sensing Object Detection},
    booktitle = {ICCV},
    year      = {2023}}

@article{li2024lsknet,
  title={LSKNet: A Foundation Lightweight Backbone for Remote Sensing},
  author={Li, Yuxuan and Li, Xiang and Dai, Yimain and Hou, Qibin and Liu, Li and Liu, Yongxiang and Cheng, Ming-Ming and Yang, Jian},
  journal={arXiv preprint arXiv:2403.11735},
  year={2024}
}

@INPROCEEDINGS{sardet100k,
	title={SARDet-100K: Towards Open-Source Benchmark and ToolKit for Large-Scale SAR Object Detection}, 
	author={Yuxuan Li and Xiang Li and Weijie Li and Qibin Hou and Li Liu and Ming-Ming Cheng and Jian Yang},
	year={2024},
	booktitle={NeurIPS},}

@INPROCEEDINGS{dota,  author={Xia, Gui-Song and Bai, Xiang and Ding, Jian and Zhu, Zhen and Belongie, Serge and Luo, Jiebo and Datcu, Mihai and Pelillo, Marcello and Zhang, Liangpei},  booktitle={CVPR},   title={{DOTA}: A Large-Scale Dataset for Object Detection in Aerial Images},   year={2018},  }

@article{orientedrcnn,
  title={Oriented r-cnn and beyond},
  author={Xie, Xingxing and Cheng, Gong and Wang, Jiabao and Li, Ke and Yao, Xiwen and Han, Junwei},
  journal={IJCV}, 
  year={2024}, 
}

@ARTICLE{droneVehicle,
  title={Drone-based RGB-Infrared Cross-Modality Vehicle Detection via Uncertainty-Aware Learning}, 
  author={Sun, Yiming and Cao, Bing and Zhu, Pengfei and Hu, Qinghua},
  journal={IEEE Transactions on Circuits and Systems for Video Technology}, 
  year={2022},
  volume={},
  number={},
  pages={1-1},
  doi={10.1109/TCSVT.2022.3168279}
}

@INPROCEEDINGS{frcnn,  author={Ren, Shaoqing and He, Kaiming and Girshick, Ross and Sun, Jian},  booktitle={NeurIPS},   title={Faster {R-CNN}: Towards Real-Time Object Detection with Region Proposal Networks},   year={2015}}

@inproceedings{liu2022convnet,
  title={A convnet for the 2020s},
  author={Liu, Zhuang and Mao, Hanzi and Wu, Chao-Yuan and Feichtenhofer, Christoph and Darrell, Trevor and Xie, Saining},
  booktitle={{CVPR}},
  year={2022}
}

@article{van,
  title={Visual Attention Network},
  author={Meng-Hao Guo and Chengrou Lu and Zheng-Ning Liu and Ming-Ming Cheng and Shiyong Hu},
  journal={Computational Visual Media},
  year={2022},
}

@article{s2anet,
	title = {Align Deep Features for Oriented Object Detection},
	journal = {TGRS},
	author = {Han, Jiaming and Ding, Jian and Li, Jie and Xia, Gui-Song},
	year = {2020},
}

@article{gfl,
  title={Generalized focal loss: Towards efficient representation learning for dense object detection},
  author={Li, Xiang and Lv, Chengqi and Wang, Wenhai and Li, Gang and Yang, Lingfeng and Yang, Jian},
  journal={TPAMI},
  year={2022},
}

@inproceedings{swin,
  title={Swin transformer: Hierarchical vision transformer using shifted windows},
  author={Liu, Ze and Lin, Yutong and Cao, Yue and Hu, Han and Wei, Yixuan and Zhang, Zheng and Lin, Stephen and Guo, Baining},
  booktitle={CVPR},
  year={2021}
}

@article{jain2024damex,
  title={DAMEX: Dataset-aware Mixture-of-Experts for visual understanding of mixture-of-datasets},
  author={Jain, Yash and Behl, Harkirat and Kira, Zsolt and Vineet, Vibhav},
  journal={NeurIPS}, 
  year={2024}
}

@inproceedings{DA,
  title={Towards universal object detection by domain attention},
  author={Wang, Xudong and Cai, Zhaowei and Gao, Dashan and Vasconcelos, Nuno},
  booktitle={CVPR}, 
  year={2019}
}

@inproceedings{unidet,
  title={Simple multi-dataset detection},
  author={Zhou, Xingyi and Koltun, Vladlen and Kr{\"a}henb{\"u}hl, Philipp},
  booktitle={CVPR}, 
  year={2022}
}

@article{riquelme2021scaling,
  title={Scaling vision with sparse mixture of experts},
  author={Riquelme, Carlos and Puigcerver, Joan and Mustafa, Basil and Neumann, Maxim and Jenatton, Rodolphe and Susano Pinto, Andr{\'e} and Keysers, Daniel and Houlsby, Neil},
  journal={NeurIPS},
  year={2021}
}

@inproceedings{GMoE,
  title={Sparse mixture-of-experts are domain generalizable learners},
  author={Li, Bo and Shen, Yifei and Yang, Jingkang and Wang, Yezhen and Ren, Jiawei and Che, Tong and Zhang, Jun and Liu, Ziwei},
  booktitle={ICLR}, 
  year={2023}
}

@inproceedings{gwd,
	title = {Rethinking Rotated Object Detection with {Gaussian} {Wasserstein} Distance Loss},
	booktitle = {ICML},
	author = {Yang, Xue and Yan, Junchi and Ming, Qi and Wang, Wentao and Zhang, Xiaopeng and Tian, Qi},
	year = {2021},
}

@inproceedings{roi_trans,
	title = {Learning {RoI} Transformer for Oriented Object Detection in Aerial Images},
	booktitle = {{CVPR}},
	author = {Ding, Jian and Xue, Nan and Long, Yang and Xia, Gui-Song and Lu, Qikai},
	year = {2019},
}

@article{dai2021attentional,
  title={Attentional local contrast networks for infrared small target detection},
  author={Dai, Yimian and Wu, Yiquan and Zhou, Fei and Barnard, Kobus},
  journal={TGRS}, 
  year={2021}, 
}

@article{devaraj2013automated,
  title={Automated geometric correction of Landsat MSS L1G imagery},
  author={Devaraj, Chabitha and Shah, Chintan A},
  journal={IEEE Geoscience and Remote Sensing Letters},
  volume={11},
  number={1},
  pages={347--351},
  year={2013},
  publisher={IEEE}
}

@article{ahamed2012tower,
  title={Tower remote-sensing system for monitoring energy crops; image acquisition and geometric corrections},
  author={Ahamed, Tofael and Tian, Lei and Jiang, Yanshui and Zhao, Bin and Liu, Hx and Ting, Kuan Chong},
  journal={Biosystems engineering},
  volume={112},
  number={2},
  pages={93--107},
  year={2012},
  publisher={Elsevier}
}

@article{liu2021multi,
  title={Multi-source remote sensing image fusion for ship target detection and recognition},
  author={Liu, Jinming and Chen, Hao and Wang, Yu},
  journal={Remote Sensing},
  volume={13},
  number={23},
  pages={4852},
  year={2021},
  publisher={MDPI}
}

@article{zhang2024hgr,
  title={HGR Correlation Pooling Fusion Framework for Recognition and Classification in Multimodal Remote Sensing Data},
  author={Zhang, Hongkang and Huang, Shao-Lun and Kuruoglu, Ercan Engin},
  journal={Remote Sensing},
  volume={16},
  number={10},
  pages={1708},
  year={2024},
  publisher={MDPI}
}

@article{zhang2024optical,
  title={Optical and Synthetic Aperture Radar Image Fusion for Ship Detection and Recognition: Current state, challenges, and future prospects},
  author={Zhang, Zenghui and Zhang, Limeng and Wu, Juanping and Guo, Weiwei},
  journal={IEEE Geoscience and Remote Sensing Magazine},
  year={2024}, 
}

@article{lin2013network,
  title={Network in network},
  author={Lin, M},
  journal={arXiv},
  year={2013}
}

@inproceedings{bozcan2020air,
  title={Au-air: A multi-modal unmanned aerial vehicle dataset for low altitude traffic surveillance},
  author={Bozcan, Ilker and Kayacan, Erdal},
  booktitle={2020 IEEE International Conference on Robotics and Automation}, 
  year={2020}, 
}

@article{avola2021low,
  title={Low-altitude aerial video surveillance via one-class SVM anomaly detection from textural features in UAV images},
  author={Avola, Danilo and Cinque, Luigi and Di Mambro, Angelo and Diko, Anxhelo and Fagioli, Alessio and Foresti, Gian Luca and Marini, Marco Raoul and Mecca, Alessio and Pannone, Daniele},
  journal={Information}, 
  year={2021}, 
}

@article{li2017earth,
  title={Earth observation brain (EOB): An intelligent earth observation system},
  author={Li, Deren and Wang, Mi and Dong, Zhipeng and Shen, Xin and Shi, Lite},
  journal={Geo-spatial information science},
  year={2017}
}

@article{anderson2017earth,
  title={Earth observation in service of the 2030 Agenda for Sustainable Development},
  author={Anderson, Katherine and Ryan, Barbara and Sonntag, William and Kavvada, Argyro and Friedl, Lawrence},
  journal={Geo-spatial Information Science},
  year={2017}
}

@inproceedings{khan2014information,
  title={Information merging in multi-UAV cooperative search},
  author={Khan, Asif and Yanmaz, Evsen and Rinner, Bernhard},
  booktitle={2014 IEEE international conference on robotics and automation}, 
  year={2014}
}

@article{jensen2016drone,
  title={Drone city--power, design and aerial mobility in the age of “smart cities”},
  author={Jensen, Ole B},
  journal={Geographica Helvetica},
  year={2016} 
}

@article{zhang2021survey,
  title={A survey on multi-task learning},
  author={Zhang, Yu and Yang, Qiang},
  journal={IEEE transactions on knowledge and data engineering},
  year={2021},
  publisher={IEEE}
}

@inproceedings{gradnorm,
  title={Gradnorm: Gradient normalization for adaptive loss balancing in deep multitask networks},
  author={Chen, Zhao and Badrinarayanan, Vijay and Lee, Chen-Yu and Rabinovich, Andrew},
  booktitle={ICML},
  year={2018}
}

@inproceedings{uncertainty,
  title={Multi-task learning using uncertainty to weigh losses for scene geometry and semantics},
  author={Kendall, Alex and Gal, Yarin and Cipolla, Roberto},
  booktitle={CVPR},
  year={2018}
}

@article{fair1m,
	title = {{FAIR1M}: A benchmark dataset for fine-grained object recognition in high-resolution remote sensing imagery},
	journal = {ISPRS},
	author = {Sun, Xian and Wang, Peijin and Yan, Zhiyuan and Xu, Feng and Wang, Ruiping and Diao, Wenhui and Chen, Jin and Li, Jihao and Feng, Yingchao and Xu, Tao and Weinmann, Martin and Hinz, Stefan and Wang, Cheng and Fu, Kun},
	year = {2022},
}

@article{zhang2021sar,
  title={SAR ship detection dataset (SSDD): Official release and comprehensive data analysis},
  author={Zhang, Tianwen and Zhang, Xiaoling and Li, Jianwei and Xu, Xiaowo and Wang, Baoyou and Zhan, Xu and Xu, Yanqin and Ke, Xiao and Zeng, Tianjiao and Su, Hao and others},
  journal={Remote Sensing},
  year={2021},
}

@MISC{Potsdam,
author = {The International Society for Photogrammetry and Remote Sensing (ISPRS)},
title = {{2D} Semantic Labeling Contest - {Potsdam}},
year = {2022},
howpublished={\url{https://www.isprs.org/education/benchmarks/UrbanSemLab/2d-sem-label-potsdam.aspx}}
}

@article{dior,
  title={Object detection in optical remote sensing images: A survey and a new benchmark},
  author={Li, Ke and Wan, Gang and Cheng, Gong and Meng, Liqiu and Han, Junwei},
  journal={ISPRS},
  year={2020}, 
}

@inproceedings{dtp,
  title={Dynamic task prioritization for multitask learning},
  author={Guo, Michelle and Haque, Albert and Huang, De-An and Yeung, Serena and Fei-Fei, Li},
  booktitle={ECCV},
  year={2018}
}

@article{pareto,
  title={Multi-task learning as multi-objective optimization},
  author={Sener, Ozan and Koltun, Vladlen},
  journal={Advances in neural information processing systems},
  year={2018}
}

@article{moe1,
  title={Adaptive mixtures of local experts},
  author={Jacobs, Robert A and Jordan, Michael I and Nowlan, Steven J and Hinton, Geoffrey E},
  journal={Neural computation},
  year={1991}
}

@article{moe2,
  title={Learning piecewise control strategies in a modular neural network architecture},
  author={Jacobs, Robert A and Jordan, Michael I},
  journal={IEEE Transactions on Systems, Man, and Cybernetics},
  year={1993}
}

@article{sparse_MoE,
  title={Outrageously large neural networks: The sparsely-gated mixture-of-experts layer},
  author={Shazeer, Noam and Mirhoseini, Azalia and Maziarz, Krzysztof and Davis, Andy and Le, Quoc and Hinton, Geoffrey and Dean, Jeff},
  journal={arXiv preprint arXiv:1701.06538},
  year={2017}
}

@inproceedings{chen2023adamv,
  title={Adamv-moe: Adaptive multi-task vision mixture-of-experts},
  author={Chen, Tianlong and Chen, Xuxi and Du, Xianzhi and Rashwan, Abdullah and Yang, Fan and Chen, Huizhong and Wang, Zhangyang and Li, Yeqing},
  booktitle={ICCV}, 
  year={2023}
}

@inproceedings{yang2024multi,
  title={Multi-Task Dense Prediction via Mixture of Low-Rank Experts},
  author={Yang, Yuqi and Jiang, Peng-Tao and Hou, Qibin and Zhang, Hao and Chen, Jinwei and Li, Bo},
  booktitle={CVPR},
  year={2024}
}

@inproceedings{xu2020universal,
  title={Universal-rcnn: Universal object detector via transferable graph r-cnn},
  author={Xu, Hang and Fang, Linpu and Liang, Xiaodan and Kang, Wenxiong and Li, Zhenguo},
  booktitle={AAAI},
  year={2020}
}

@article{kapidis2021multi,
  title={Multi-dataset, multitask learning of egocentric vision tasks},
  author={Kapidis, Georgios and Poppe, Ronald and Veltkamp, Remco C},
  journal={TPAMI},
  year={2021}
}

@inproceedings{zhao2020object,
  title={Object detection with a unified label space from multiple datasets},
  author={Zhao, Xiangyun and Schulter, Samuel and Sharma, Gaurav and Tsai, Yi-Hsuan and Chandraker, Manmohan and Wu, Ying},
  booktitle={2020},
  year={2020}
}

@article{yan2020learning,
  title={Learning from multiple datasets with heterogeneous and partial labels for universal lesion detection in CT},
  author={Yan, Ke and Cai, Jinzheng and Zheng, Youjing and Harrison, Adam P and Jin, Dakai and Tang, Youbao and Tang, Yuxing and Huang, Lingyun and Xiao, Jing and Lu, Le},
  journal={IEEE Transactions on Medical Imaging},
  year={2020}
}

@inproceedings{yang2019hierarchical,
  title={Hierarchical deep stereo matching on high-resolution images},
  author={Yang, Gengshan and Manela, Joshua and Happold, Michael and Ramanan, Deva},
  booktitle={CVPR}, 
  year={2019}
}

@inproceedings{SEnet,
  title={Squeeze-and-excitation networks},
  author={Hu, Jie and Shen, Li and Sun, Gang},
  booktitle={CVPR}, 
  year={2018}
}

@inproceedings{liu_improving_2020,
    title = {Improving Convolutional Networks With Self-Calibrated Convolutions},
    booktitle = {{CVPR}},
    author = {Liu, Jiang-Jiang and Hou, Qibin and Cheng, Ming-Ming and Wang, Changhu and Feng, Jiashi},
    year = {2020}
}

@article{nakano2021cross,
  title={Cross-task consistency learning framework for multi-task learning},
  author={Nakano, Akihiro and Chen, Shi and Demachi, Kazuyuki},
  journal={arXiv},
  year={2021}
}

@article{huang2024low,
  title={Low-Altitude Intelligent Transportation: system architecture, infrastructure, and key technologies},
  author={Huang, Changqing and Fang, Shifeng and Wu, Hua and Wang, Yong and Yang, Yichen},
  journal={Journal of Industrial Information Integration},
  pages={100694},
  year={2024},
  publisher={Elsevier}
}

@article{jiang20236g,
  title={6G Non-Terrestrial networks enabled low-altitude economy: Opportunities and challenges},
  author={Jiang, Yihang and Li, Xiaoyang and Zhu, Guangxu and Li, Hang and Deng, Jing and Shi, Qingjiang},
  journal={arXiv preprint arXiv:2311.09047},
  year={2023}
}

@inproceedings{elangovan2016medical,
  title={Medical imaging modalities: a survey},
  author={Elangovan, Aarthipoornima and Jeyaseelan, Thangaraja},
  booktitle={International Conference on emerging trends in engineering, technology and science},
  year={2016}
}

@article{feng2020deep,
  title={Deep multi-modal object detection and semantic segmentation for autonomous driving: Datasets, methods, and challenges},
  author={Feng, Di and Haase-Sch{\"u}tz, Christian and Rosenbaum, Lars and Hertlein, Heinz and Glaeser, Claudius and Timm, Fabian and Wiesbeck, Werner and Dietmayer, Klaus},
  journal={IEEE Transactions on Intelligent Transportation Systems},
  year={2020},
}

@INPROCEEDINGS{retina,  author={Lin, Tsung-Yi and Goyal, Priya and Girshick, Ross and He, Kaiming and Dollár, Piotr},  booktitle={ICCV},   title={Focal Loss for Dense Object Detection},   year={2017}}

@INPROCEEDINGS{cascade,  author={Cai, Zhaowei and Vasconcelos, Nuno},  booktitle={CVPR},   title={Cascade {R-CNN}: Delving Into High Quality Object Detection},   year={2018},  volume={},  number={}}

@article{li2024predicting,
  title = {Predicting gradient is better: Exploring self-supervised learning for SAR ATR with a joint-embedding predictive architecture},
  journal = {ISPRS Journal o },
  year = {2024},
  author = {Li, Weijie and Yang, Wei and Liu, Tianpeng and Hou, Yuenan and Li, Yuxuan and Liu, Zhen and Liu, Yongxiang and Liu, Li},
}

@article{dai2024denodet,
  title={DenoDet: Attention as Deformable Multi-Subspace Feature Denoising for Target Detection in SAR Images},
  author={Dai, Yimian and Zou, Minrui and Li, Yuxuan and Li, Xiang and Ni, Kang and Yang, Jian},
  journal={arXiv},
  year={2024}
}

@article{msar,
  title={CRTransSar: A visual transformer based on contextual joint representation learning for SAR ship detection},
  author={Xia, Runfan and Chen, Jie and Huang, Zhixiang and Wan, Huiyao and Wu, Bocai and Sun, Long and Yao, Baidong and Xiang, Haibing and Xing, Mengdao},
  journal={Remote Sensing},
  year={2022},
}

@article{SIVED,
  title={SIVED: A SAR Image Dataset for Vehicle Detection Based on Rotatable Bounding Box},
  author={Lin, Xin and Zhang, Bo and Wu, Fan and Wang, Chao and Yang, Yali and Chen, Huiqin},
  journal={Remote Sensing},
  year={2023},
}

@article{sadd,
  title={SEFEPNet: Scale expansion and feature enhancement pyramid network for SAR aircraft detection with small sample dataset},
  author={Zhang, Peng and Xu, Hao and Tian, Tian and Gao, Peng and Li, Linfeng and Zhao, Tianming and Zhang, Nan and Tian, Jinwen},
  journal={IEEE Journal of Selected Topics in Applied Earth Observations and Remote Sensing},
  year={2022},
}

@article{ogsod,
  title={Category-oriented Localization Distillation for SAR Object Detection and A Unified Benchmark},
  author={Wang, Chao and Ruan, Rui and Zhao, Zhicheng and Li, Chenglong and Tang, Jin},
  journal={IEEE Transactions on Geoscience and Remote Sensing},
  year={2023},
  publisher={IEEE}
}

@article{wei2020hrsid,
  title={HRSID: A high-resolution SAR images dataset for ship detection and instance segmentation},
  author={Wei, Shunjun and Zeng, Xiangfeng and Qu, Qizhe and Wang, Mou and Su, Hao and Shi, Jun},
  journal={IEEE Access},
  year={2020},
}

@article{wang2022pvt,
  title={Pvt v2: Improved baselines with pyramid vision transformer},
  author={Wang, Wenhai and Xie, Enze and Li, Xiang and Fan, Deng-Ping and Song, Kaitao and Liang, Ding and Lu, Tong and Luo, Ping and Shao, Ling},
  journal={Computational Visual Media},
  year={2022}
}

@article{zhang2025unichange,
  title={UniChange: Unifying Change Detection with Multimodal Large Language Model},
  author={Zhang, Xu and Li, Danyang and Dong, Xiaohang and Wu, Tianhao and Yu, Hualong and Wang, Jianye and Li, Qicheng and Li, Xiang},
  journal={arXiv preprint arXiv:2511.02607},
  year={2025}
}

@article{Li_2025_ViTP,
  title={Visual Instruction Pretraining for Domain-Specific Foundation Models},
  author={Li, Yuxuan and Zhang, Yicheng and Tang, Wenhao and Dai, Yimian and Cheng, Ming-Ming and Li, Xiang and Yang, Jian},
  journal={arXiv},
  year={2025}
}

@article{denodetv2,
  title={DenoDet V2: Phase-Amplitude Cross Denoising for SAR Object Detection},
  author={Ni, Kang and Zou, Minrui and Li, Yuxuan and Li, Xiang and Guo, Kehua and Cheng, Ming-Ming and Dai, Yimian},
  journal={AAAI},
  year={2025}
}

@article{yuan2025strip,
  title={Strip R-CNN: Large Strip Convolution for Remote Sensing Object Detection},
  author={Yuan, Xinbin and Zheng, ZhaoHui and Li, Yuxuan and Liu, Xialei and Liu, Li and Li, Xiang and Hou, Qibin and Cheng, Ming-Ming},
  journal={arXiv preprint arXiv:2501.03775},
  year={2025}
}

\newpage

\appendix

\section*{A. Datasets}
The \textbf{SARDet-100K}~\cite{sardet100k} dataset is a SAR object detection dataset containing six object categories: Aircraft, Ship, Car, Bridge, Tank, and Harbor.
The dataset consists of 94,493 training images with 198,747 instances, and 11,613 testing images with 24,023 instances. 
All annotations are provided as horizontal bounding boxes (HBB). 
\textbf{DOTA}~\cite{dota} is an optical aerial object detection dataset that includes 15 categories. After splitting each image into 800$\times$800 patches with a 400-pixel overlap, the dataset yields 25,028 training images containing 337,728 instances, and 17,041 testing images with 95,380 instances. All annotations are in the form of oriented bounding boxes (OBB). To avoid the severe dataset imbalance in the merged SOI-Det dataset, we use only a subset of SARDet-100K. For more details please refer to the Supplementary Material.
\textbf{DroneVehicle}~\cite{droneVehicle} is an infrared vehicle detection dataset with 5 categories: car, truck, bus, van, and freight car. The images are sized at 640$\times$512 pixels. The dataset consists of 17,990 training images with 316,411 instances, and 8,980 testing images with 159,616 instances. Similar to DOTA-v1.0, all annotations are in the form of OBB.

To avoid the severe dataset imbalance in the merged SOI-Det dataset, we use only a subset of SARDet-100K: HRSID~\cite{wei2020hrsid}, MSAR~\cite{msar}, SADD~\cite{sadd}, OGSOD~\cite{ogsod}, and SIVED~\cite{SIVED}. This subset of SARDet-100K includes 47,097 training images with 125,462 annotated instances and 4,481 testing images with 12,566 instances. During training, each batch is sampled uniformly across the three datasets—SARDet-100K, DOTA, and DroneVehicle—in a 2:1:1 ratio, ensuring that each dataset is cycled through approximately once every 20K iterations.

For each dataset, we evaluate using the mean Average Precision at IoU thresholds 0.5 (@50), at IoU thresholds 0.75 (@75), and at IoU thresholds from 0.5 to 0.95 (mAP). Additionally, we report the overall mAP across the three datasets to assess overall performance.

\begin{table*}[!t]
  \centering
\begin{tabular}{cccc|cc|ccc}
Stage 1 & Stage 2 & Stage 3 & Stage 4 & FLOPs & ~~\#P~~ & \textbf{mAP} & @50 & @75 \\ \Xhline{1pt}
\textcolor{gray}{None} & \textcolor{gray}{None} &\textcolor{gray}{None} & \textcolor{gray}{None}  &  403G & 66M   &  48.51  &  79.70   &  51.78  \\
\hline
\textcolor{gray}{None} & \textcolor{gray}{None}  & \textcolor{gray}{None}  & Even     &  422G   & 132M    &    48.85{+(0.34)} &  80.07{(+0.37)}   &  51.86{+(0.08)}  \\
\textcolor{gray}{None} & \textcolor{gray}{None} & Even     & Even     &   469G    &  174M   &   49.31{+(0.80)}  &  80.26{+(0.56)}   &  52.84{+(1.06)}   \\
\textcolor{gray}{None}  & Even  & Even  & Even  &  487G & 178M &  \textbf{49.53{+(1.02)}} & \textbf{80.47{+(0.77)}} & \textbf{53.06{+(1.28)} }   \\
Even    & Even    & Even     & Even     &  506G     & 179M    &   49.47{+(0.96)}  &  80.33{+(0.63)}   &  52.98{+(1.20)}   \\
All     & All     & All     & All     &   572G  &  249M   &  49.30 {+(0.79)}    &  80.23{+(0.53)}    & 53.03{+(1.25)}    \\
\end{tabular}
\caption{Experiments on spatial MoE with different MoE layer positions. ``None'': no MoE layers, ``Even'': MoE added to even-indexed layers, and ``All'': MoE added to all layers within the stage. Each MoE layer comprises 8 experts with a top-2 selection. Selectively incorporating MoE layers in the even-indexed layers of the last three stages enhances model performance.} 
\label{tab:MoE_blks}
\end{table*}

\section*{B. Implementation Details}
All models are fine-tuned on their respective training sets. For SARDet-100K and DroneVehicle, models are evaluated on the corresponding test sets, while for the DOTA dataset, evaluation is conducted on the validation set.
For individual dataset training, models are trained for 12 epochs using the AdamW optimizer. 

In multi-modal joint training, we ensure that the total number of iterations matches the combined iterations of individual dataset training for fairness. Each batch employs uniform sampling from the three datasets (SARDet-100K, DOTA, and DroneVehicle) with a ratio of 2:1:1, ensuring that all datasets are cycled through approximately once every 20K iterations. Following previous network designs \cite{DA,unidet}, we share the backbone network and use separate task heads for different datasets and tasks. 
{Specifically, after feature extraction by the backbone network, features from SARDet-100K images are passed to the GFL head (due to its superior performance on horizontal SAR object detection ~\cite{sardet100k}), while those from DOTA and DroneVehicle are passed to two individual O-RCNN~\cite{orientedrcnn} heads (due to the high-performance of O-RCNN on oriented object detection).} 

In the main results and ablation studies, ConvNext-T is used as the default backbone unless otherwise specified. We run every experiment once. The initial learning rate is set to 0.0001, with a weight decay of 0.05. Model training is conducted using 8 RTX 3090 GPUs, with a batch size of 4 per GPU. All FLOPs reported in this paper are calculated using an 800×800 image input.

\section*{C. Grid-level sparse MoE analysis}

\textbf{MoE Layer Positions.}
We conduct an ablation study to assess the impact of incorporating MoE layers at different stages of the ConvNext backbone. As shown in Table~\ref{tab:MoE_blks}, selectively adding MoE layers in the last three stages enhances model performance, resulting in a 1.02\% mAP improvement with a minor increase in computational cost. This enhancement likely results from the richer semantic information in deeper stages, allowing experts to specialize more effectively. Conversely, abusing MoE layers does not lead to optimal performance, indicating that too many experts may introduce optimization challenges, as also highlighted in recent studies~\cite{riquelme2021scaling,GMoE}.

\begin{figure*}[!ht]
  \centering
  \includegraphics[width=0.76\linewidth]{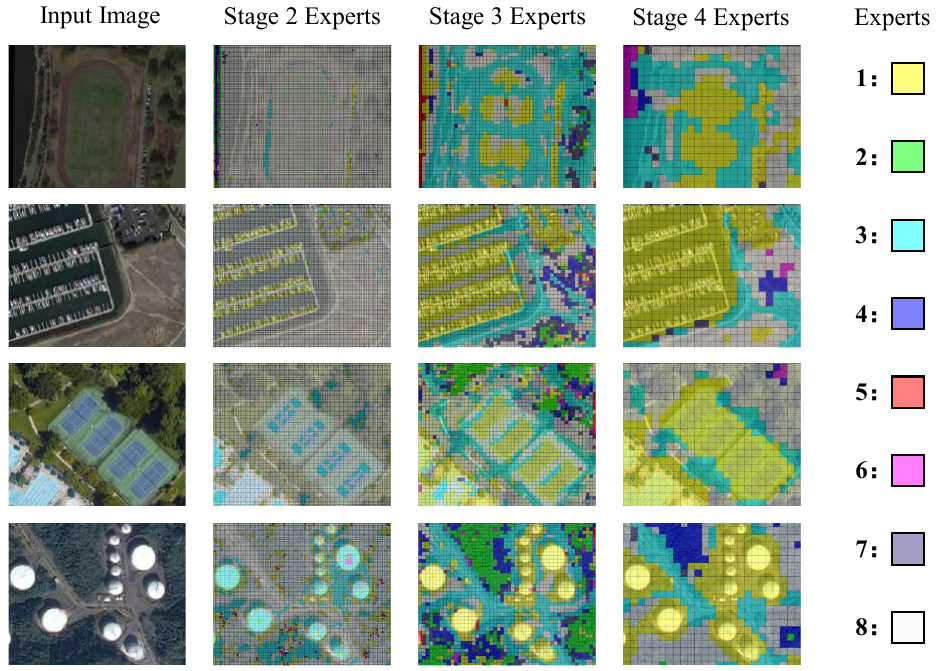}
  \caption{Visualization of grid-level expert activation across the last three stages of a well-tuned ConvNext-T backbone on DOTA-v1.0 images. Each square grid represents the receptive field at a given stage, with different colours indicating the local grid areas processed by distinct experts. The top-1 selected experts for each grid are shown.}
  \label{fig:sm_experts2}
\end{figure*}

\begin{figure*}[!ht]
  \centering
  \includegraphics[width=0.76\linewidth]{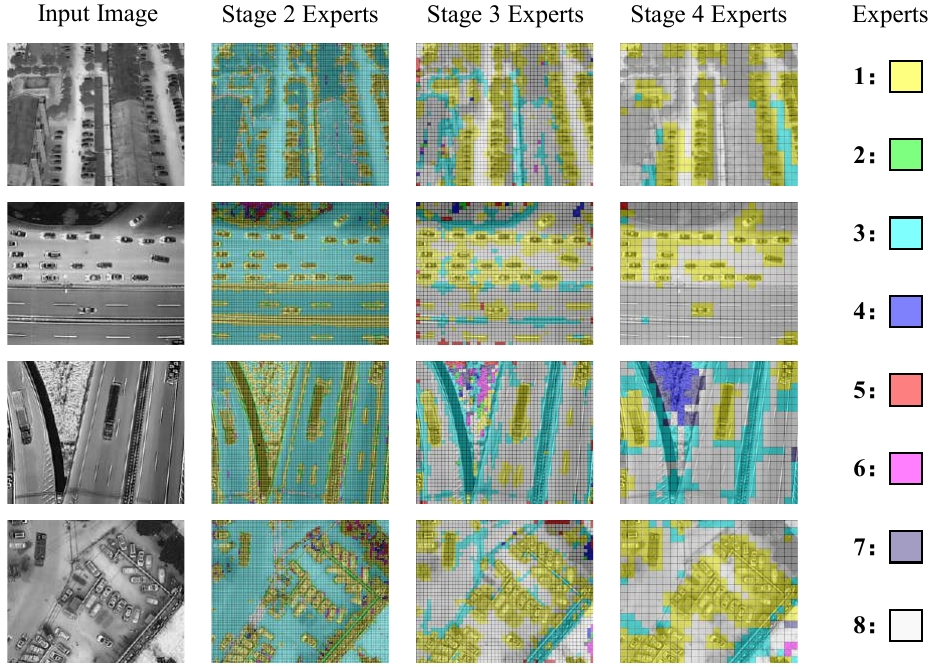}
  \caption{Visualization of grid-level expert activation across the last three stages of a well-tuned ConvNext-T backbone on DroneVehicle images. Each square grid represents the receptive field at a given stage, with different colours indicating the local grid areas processed by distinct experts. The top-1 selected experts for each grid are shown.}
  \label{fig:sm_experts1}
\end{figure*}

\begin{figure*}[!hb]
  \centering
  \includegraphics[width=0.76\linewidth]{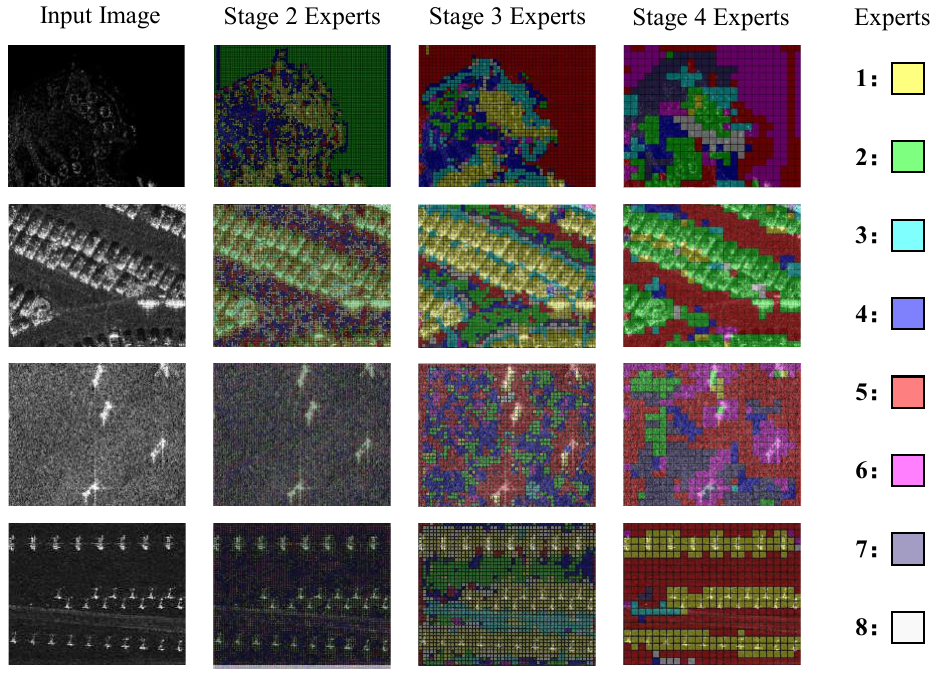}
  \caption{Visualization of grid-level expert activation across the last three stages of a well-tuned ConvNext-T backbone on SARDet-100K images. Each square grid represents the receptive field at a given stage, with different colours indicating the local grid areas processed by distinct experts. The top-1 selected experts for each grid are shown.}
  \label{fig:sm_experts3}
\end{figure*}

\textbf{Grid-level Expert Activation.} Visualization of grid expert activation across the last three stages of a well-tuned ConvNext-T backbone on DOTA, DroneVehicle and SARDet-100K images are given in Figure~\ref{fig:sm_experts2}, \ref{fig:sm_experts1} and \ref{fig:sm_experts3}. Each square grid represents the receptive field at a given stage, with different colours indicating the local grid areas processed by
distinct experts. The top-1 selected experts for each grid are shown. 

\begin{figure*}[!ht]
  \centering
  \includegraphics[width=0.8\linewidth]{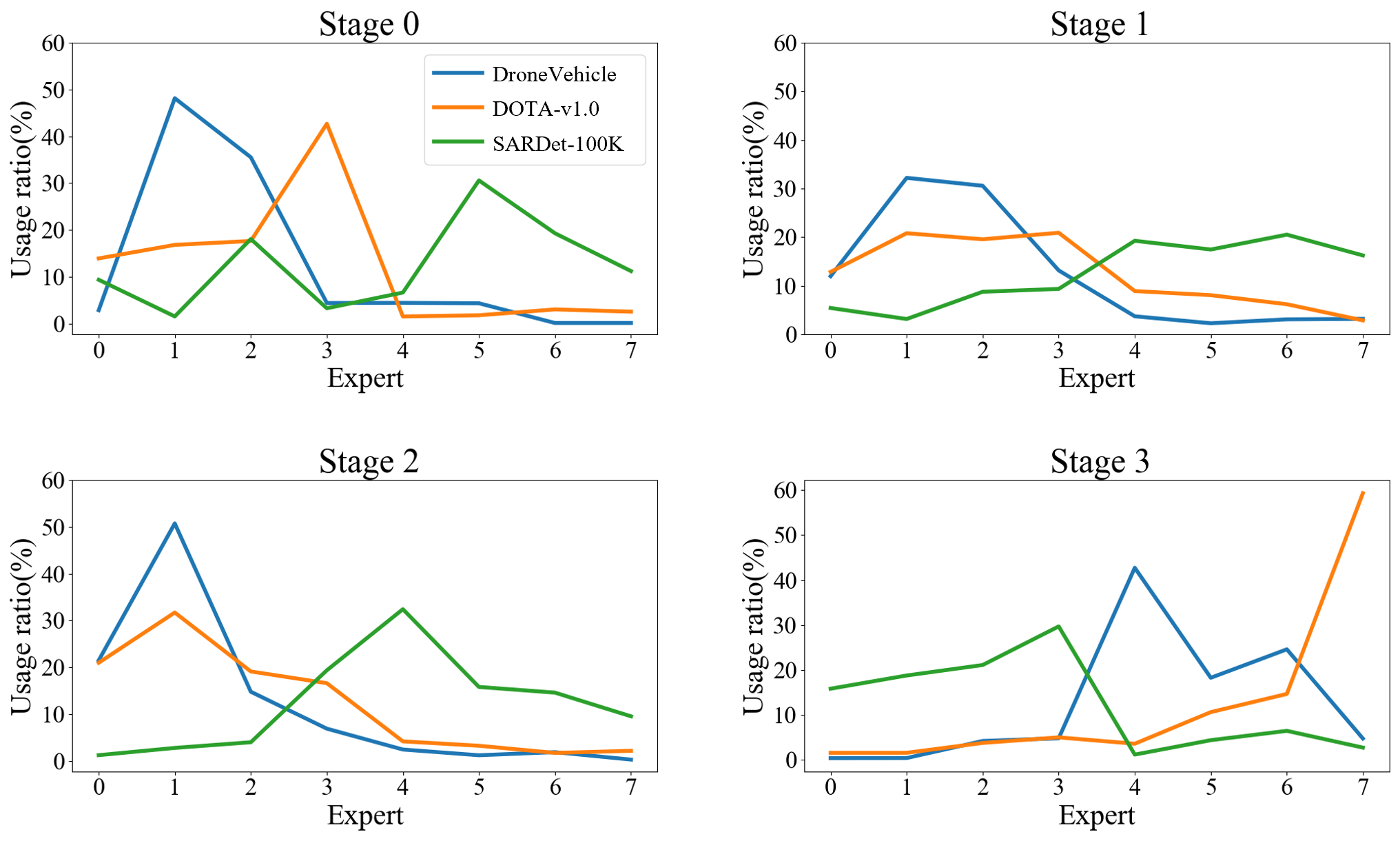}
  \caption{Expert participation statistics across SARDet-100K, DOTA, and DroneVehicle datasets. Some experts contribute to shared representation learning across all modalities, while others specialize in distinct patterns, activating only specific modalities.}
  \label{fig:expert_dataset_stat}
\end{figure*}

In addition to the visualization, we further analyze the behaviour of sparse MoE expert selection at the dataset scale. We pass all test images from SARDet-100K, DOTA, and DroneVehicle datasets through the well-trained SM3Det model and gather statistics on expert participation. Specifically, the participation of an expert in a given MoE layer is quantified by the softmax probability of the gate function in main paper Eq.~(2). The statistical results, presented in Figure~\ref{fig:expert_dataset_stat}, These findings validate the advantage of our sparse MoE design in addressing the M2Det task. The expert selection patterns demonstrate that some experts contribute to shared representation learning across all three modalities, while others specialize in distinct patterns, activating only specific modalities. This balance supports both joint and independent representation learning.
Notably, experts activated by SAR images typically show low activation in the other two modalities, indicating that SAR images utilize a distinct set of experts. In contrast, Infrared and RGB images share several highly activated experts, reflecting greater overlap in their representations. This observation aligns with the conventional understanding that SAR imagery embodies unique concepts and characteristics distinct from those in other modalities.


\section*{D. Detailed Experiment Results}

\subsection*{D.1 Parameter Efficiency.}

\begin{table}[!h]

\centering
\begin{tabular}{cc|cc}
\multicolumn{2}{c|}{Size(M)} & \multicolumn{2}{c}{mAP} \\
3 models & SM3Det & 3 models & SM3Det \\
\hline
192 \textcolor[RGB]{112,173,71}{S} & 178 \textcolor[RGB]{91,155,213}{T} & 49.17 & \textbf{50.20} \\
309 \textcolor[RGB]{237,125,49}{B} & 275 \textcolor[RGB]{112,173,71}{S} & 50.18 & \textbf{50.28} \\
636 \textcolor[RGB]{255,0,0}{L} & 459 \textcolor[RGB]{237,125,49}{B} & 50.5 & \textbf{51.33} \\
- & 770 \textcolor[RGB]{255,0,0}{L} & - & \textbf{52.16}
\end{tabular}
\caption{\textcolor[RGB]{91,155,213}{T}, \textcolor[RGB]{112,173,71}{S}, \textcolor[RGB]{237,125,49}{B}, \textcolor[RGB]{255,0,0}{L}: Backbone with ConvNext size of \textcolor[RGB]{91,155,213}{Tiny}, \textcolor[RGB]{112,173,71}{Small}, \textcolor[RGB]{237,125,49}{Base} and \textcolor[RGB]{255,0,0}{Large}. SM3Det consistently outperforms baseline models, even with fewer parameters. Its advantage increases with larger models. This demonstrates SM3Det’s robustness and effectiveness.}
\label{tab:efficiency}
\end{table}

The results in Table~\ref{tab:efficiency} demonstrate SM3Det’s superior performance and parameter efficiency. Notably, SM3Det achieves performance gains even when compared to larger baseline models, underscoring the critical role of its consistency and synchronization optimization strategy rather than mere parameter scaling. For instance, SM3Det-Tiny (178M, T) outperforms the baseline Small model (192M, S) with a 1.03\% mAP improvement (50.20 vs. 49.17) despite having 7.3\% fewer parameters, highlighting that performance gains stem from effective multi-modal learning rather than parameter inflation. Similarly, SM3Det-Small (275M, S) surpasses the baseline Base model (309M, B) by 0.1\% mAP (50.28 vs. 50.18) with 11\% fewer parameters, further validating that SM3Det’s advantages are not driven by model scale but by its ability to harmonize multi-modal and multi-task optimization.

The scalability advantage becomes more pronounced in larger configurations: SM3Det-Base (459M, B) achieves a 0.83\% mAP gain (51.33 vs. 50.5) over the baseline Large model (636M, L), using 27.8\% fewer parameters. Remarkably, SM3Det-Large (770M, L) can easily keep scaling up, it sets a new state-of-the-art (52.16 mAP) without a direct baseline counterpart, demonstrating its unique capacity to leverage cross-modal synergies. These results confirm that the grid-level sparse MoE backbone, coupled with dynamic learning rate synchronization, mitigates parameter conflicts and ensures balanced learning across modalities and tasks. Crucially, even in cross-scale comparisons (e.g., SM3Det-Small vs. baseline-Base), SM3Det maintains superior efficiency and accuracy, proving that its performance stems from systematic optimization rather than brute-force parameter scaling. This aligns with the core innovation of SM3Det: unifying multi-modal learning while preserving task-specific discriminability through adaptive optimization.

\begin{table*}[!t]
  \centering
\begin{tabular}{c|cc|cccc}
MoE Cfg &  FLOPs &  \#P &  Test & \textbf{mAP} &  @50 &  @75 \\ \Xhline{1pt}
   &   &   &  \cellcolor[HTML]{EFEFEF}Overall &   \cellcolor[HTML]{EFEFEF} 48.51 &   \cellcolor[HTML]{EFEFEF} 79.70 &   \cellcolor[HTML]{EFEFEF} 51.78 \\ \cline{4-7} 
   &   &   &   SARDet-100K & 56.99  & 87.32  & 61.41  \\
   &   &   &   DOTA & 45.59 & 77.86  & 47.01  \\
  \multirow{-4}{*}{\begin{tabular}[c]{@{}c@{}} w/o MoE \end{tabular}}  &  \multirow{-4}{*}{403G} &  \multirow{-4}{*}{66M} &  DroneVehicle & 47.09  & 76.10  & 54.53  \\ \Xhline{1pt}
   &   &   &  \cellcolor[HTML]{EFEFEF}Overall & 48.94 \cellcolor[HTML]{EFEFEF} & 80.25\cellcolor[HTML]{EFEFEF} &  52.01\cellcolor[HTML]{EFEFEF} \\ \cline{4-7} 
   &   &   &  SARDet-100K & 57.81  & 88.33 & 61.96  \\
   &   &   &  DOTA & 45.89  & 78.34  & 47.15  \\
  \multirow{-4}{*}{\begin{tabular}[c]{@{}c@{}}Experts: 2 \\ Top: ~~~~~~2 \end{tabular}} &  \multirow{-4}{*}{469G} &  \multirow{-4}{*}{82M} &  DroneVehicle & 47.42 & 76.28  & 54.65  \\ \Xhline{1pt}
   &   &   &  \cellcolor[HTML]{EFEFEF}Overall & 49.11 \cellcolor[HTML]{EFEFEF} &  80.10\cellcolor[HTML]{EFEFEF} &
  52.13\cellcolor[HTML]{EFEFEF} \\ \cline{4-7} 
   &   &   &  SARDet-100K & 58.52  & 88.39 & 62.47  \\
   &   &   &  DOTA & 45.84  & 77.99 & 46.98  \\
  \multirow{-4}{*}{\begin{tabular}[c]{@{}c@{}}Experts: 4 \\ Top: ~~~~~~2 \end{tabular}} &  \multirow{-4}{*}{469G} &  \multirow{-4}{*}{113M} &  DroneVehicle & 47.64  & 76.49  & 55.17  \\ \Xhline{1pt}
   &   &   &  \cellcolor[HTML]{EFEFEF}Overall &  49.11\cellcolor[HTML]{EFEFEF} & 79.74  \cellcolor[HTML]{EFEFEF} & 52.76\cellcolor[HTML]{EFEFEF} \\ \cline{4-7} 
   &   &   &  SARDet-100K &  59.16 &  88.89 & 64.10  \\
   &   &   &  DOTA & 45.42  &  76.94 &  47.15 \\
  \multirow{-4}{*}{\begin{tabular}[c]{@{}c@{}}Experts: 6 \\ Top: ~~~~~~2 \end{tabular}} &  \multirow{-4}{*}{469G} &  \multirow{-4}{*}{143M} &  DroneVehicle & 48.14  &  77.14 & 56.00 \\ \Xhline{1pt}
   &   &   &  \cellcolor[HTML]{EFEFEF}Overall &  49.31\cellcolor[HTML]{EFEFEF} & 80.26 \cellcolor[HTML]{EFEFEF} & 52.84 \cellcolor[HTML]{EFEFEF} \\ \cline{4-7} 
   &   &   &  SARDet-100K &  58.99 & 88.89  & 63.87  \\
   &   &   &  DOTA & 45.69  &  77.64 &  47.13 \\
  \multirow{-4}{*}{\begin{tabular}[c]{@{}c@{}}Experts: 8 \\ Top: ~~~~~~2 \end{tabular}} &  \multirow{-4}{*}{469G} &  \multirow{-4}{*}{174M} &  DroneVehicle &  48.53 &  77.74 & 56.71 \\  \Xhline{1pt}
   &   &   &  \cellcolor[HTML]{EFEFEF}Overall & 49.24 \cellcolor[HTML]{EFEFEF} & 80.18 \cellcolor[HTML]{EFEFEF} &  52.79 \cellcolor[HTML]{EFEFEF} \\ \cline{4-7} 
   &   &   &  SARDet-100K &  59.24 &  89.11 & 63.78  \\
   &   &   &  DOTA & 45.65  & 77.54  & 47.27  \\
  \multirow{-4}{*}{\begin{tabular}[c]{@{}c@{}}Experts: 10 \\ Top: ~~~~~~2 \end{tabular}} &  \multirow{-4}{*}{469G} &  \multirow{-4}{*}{205M} &  DroneVehicle &  48.02 & 77.37  & 56.17 \\  \hline \hline
   &   &   &  \cellcolor[HTML]{EFEFEF}Overall & 49.05 \cellcolor[HTML]{EFEFEF} &   \cellcolor[HTML]{EFEFEF} 79.72 &  52.30 \cellcolor[HTML]{EFEFEF} \\ \cline{4-7} 
   &   &   &  SARDet-100K &  59.10 &  88.60 & 63.97  \\
   &   &   &  DOTA & 45.23  & 76.88  & 46.31 \\
  \multirow{-4}{*}{\begin{tabular}[c]{@{}c@{}}Experts: 8 \\ Top: ~~~~~~1 \end{tabular}} &  \multirow{-4}{*}{403G} &  \multirow{-4}{*}{174M} &  DroneVehicle &  48.44 & 77.60 & 56.28 \\ \Xhline{1pt}
   &   &   &  \cellcolor[HTML]{EFEFEF}Overall &  49.31\cellcolor[HTML]{EFEFEF} & 80.26 \cellcolor[HTML]{EFEFEF} & 52.84 \cellcolor[HTML]{EFEFEF} \\ \cline{4-7} 
   &   &   &  SARDet-100K &  58.99 & 88.89  & 63.87  \\
   &   &   &  DOTA & 45.69  &  77.64 &  47.13 \\
  \multirow{-4}{*}{\begin{tabular}[c]{@{}c@{}}Experts: 8 \\ Top: ~~~~~~2 \end{tabular}} &  \multirow{-4}{*}{469G} &  \multirow{-4}{*}{174M} &  DroneVehicle &  48.53 &  77.74 & 56.71 \\  \Xhline{1pt}
   &   &   &  \cellcolor[HTML]{EFEFEF}Overall & 49.13 \cellcolor[HTML]{EFEFEF} & 79.98 \cellcolor[HTML]{EFEFEF} &  52.77 \cellcolor[HTML]{EFEFEF} \\ \cline{4-7} 
   &   &   &  SARDet-100K &  59.18 &  88.98 & 63.65  \\
   &   &   &  DOTA & 45.43  & 77.32  & 47.33  \\
  \multirow{-4}{*}{\begin{tabular}[c]{@{}c@{}}Experts: 8 \\ Top: ~~~~~~3 \end{tabular}} &  \multirow{-4}{*}{531G} &  \multirow{-4}{*}{174M} &  DroneVehicle &  48.16 & 77.15  & 56.05 \\ \hline \hline
   &   &   &  \cellcolor[HTML]{EFEFEF}Overall &  48.49 \cellcolor[HTML]{EFEFEF} &   79.60\cellcolor[HTML]{EFEFEF} &   51.51\cellcolor[HTML]{EFEFEF} \\ \cline{4-7} 
   &   &   &  SARDet-100K &  56.88  &  87.17  & 60.49  \\
   &   &   &  DOTA &   45.56 &  77.74  &  46.94  \\
  \multirow{-4}{*}{\begin{tabular}[c]{@{}c@{}} Image-level \\ Experts: 3 \\ Top: ~~~~~~1 \end{tabular}} &  \multirow{-4}{*}{ 403G} &  \multirow{-4}{*}{ 98M} &  DroneVehicle & 47.21  &  76.07  & 54.45  \\  \Xhline{1pt}
   &   &   &  \cellcolor[HTML]{EFEFEF}Overall &  48.60 \cellcolor[HTML]{EFEFEF} &   \cellcolor[HTML]{EFEFEF} 79.67 &  52.06  \cellcolor[HTML]{EFEFEF} \\ \cline{4-7} 
   &   &   &  SARDet-100K &  56.51  & 87.23  & 61.00   \\
   &   &   &  DOTA &  45.80 &  77.90  & 47.51   \\
  \multirow{-4}{*}{\begin{tabular}[c]{@{}c@{}}Grid level \\ Experts: 3 \\ Top: ~~~~~~1 \end{tabular}} &  \multirow{-4}{*}{ 403G} &  \multirow{-4}{*}{ 98M} &  DroneVehicle &  47.50 &  75.93  & 54.96

\end{tabular}
  \caption{Experiments on grid-level MoE with varying numbers of experts and top-K selection configurations. Experts are applied only to the even-indexed layers of the last two stages for validation efficiency. }
  \label{tab:MoE_experts_detailed}
\end{table*}

\subsection*{D.2 Ablation study on MoE configurations}

We investigate the impact of different configurations of the sparse MoE architecture on model performance. Specifically, we analyze variations in the number of experts and the Top-K value, which determines the number of experts activated for each input. The detailed results are given in Table~\ref{tab:MoE_experts_detailed}.

As the number of experts increases from 2 to 10 with a fixed Top-K value of 2, there is a consistent improvement in overall mAP scores until reaching 8 experts. This suggests that up to a certain point, adding more experts allows the model to better capture diverse patterns in the data, leading to improved detection performance across multiple datasets. When the number of experts is increased to 10, the overall mAP slightly decreases to 49.24. This indicates that while adding more experts can enhance model capacity, there may be diminishing returns or even a negative impact when the number of experts exceeds the capacity of the training data to sufficiently train them all.

When examining the impact of the Top-K value, it is clear that setting Top-K to 2 generally provides a good balance between performance and computational demand. 

For example, with 8 experts and a Top-K value of 2, the overall mAP is 49.31, which is higher than both the Top-K values of 1 and 3. A Top-K value of 1, while less computationally intensive, results in a slightly lower mAP of 49.05, indicating that activating only one expert per input may limit the model's capacity to leverage the diverse expertise available. Conversely, increasing the Top-K value to 3, while introducing more computational demands, does not improve performance, as the overall mAP drops to 49.13. 

This suggests that activating only one expert per input may limit the model's capacity to leverage the diverse expertise available, while activating too many experts for a single input may lead to over-complexity without corresponding benefits, potentially causing interference between experts or inefficient use of computational resources.

In summary, the ablation study reveals that the optimal configuration for this sparse MoE architecture in terms of balancing performance and computational efficiency is 8 experts with a Top-K value of 2. This configuration maximizes the model's ability to learn from diverse inputs without introducing unnecessary complexity or overfitting. 


\begin{table*}[t]
  \centering
\begin{tabular}{cc|cccc}
T &  b  &  Test & \textbf{mAP} &  @50 &  @75 \\ \Xhline{1pt}
   &    &  \cellcolor[HTML]{EFEFEF}Overall &   \cellcolor[HTML]{EFEFEF}49.47 &   \cellcolor[HTML]{EFEFEF}80.33 &   \cellcolor[HTML]{EFEFEF}52.98 \\ \cline{3-6} 
   &    &  SARDet-100K & 58.97  & 88.82 & 63.40  \\
   &    &  DOTA & 46.02 & 77.84  & 47.73  \\
  \multirow{-4}{*}{\hspace{4pt}w/o}  &  \multirow{-4}{*}{\hspace{-10pt}DSO} &  DroneVehicle & 48.41  & 77.62  & 56.23  \\ \Xhline{1pt}
   &    &  \cellcolor[HTML]{EFEFEF}Overall &  \cellcolor[HTML]{EFEFEF} 50.14 &  \cellcolor[HTML]{EFEFEF} 80.61 &  \cellcolor[HTML]{EFEFEF} 53.81 \\ \cline{3-6} 
   &   &  SARDet-100K & 60.86  & 90.02 & 66.01 \\
   &   &  DOTA & 46.23  & 77.68  & 47.81  \\
  \multirow{-4}{*}{3}  &  \multirow{-4}{*}{0.3} &  DroneVehicle & 49.03  & 77.99  & 56.90  \\ \Xhline{1pt}
  &    &  \cellcolor[HTML]{EFEFEF}Overall &   \cellcolor[HTML]{EFEFEF}50.20 &   \cellcolor[HTML]{EFEFEF}80.68  &   \cellcolor[HTML]{EFEFEF}53.79 \\ \cline{3-6} 
   &    &  SARDet-100K & 60.64  & 89.94 & 65.06  \\
   &    &  DOTA & 46.47 & 77.88  & 48.24  \\
  \multirow{-4}{*}{3}  &  \multirow{-4}{*}{0.4} &  DroneVehicle & 48.87  & 77.99  & 56.90  \\ \Xhline{1pt}
   &   &  \cellcolor[HTML]{EFEFEF}Overall &  \cellcolor[HTML]{EFEFEF} 50.07 &  \cellcolor[HTML]{EFEFEF} 80.66 & \cellcolor[HTML]{EFEFEF} 54.00 \\ \cline{3-6} 
   &   &  SARDet-100K & 60.77  & 90.04  & 65.74  \\
   &   &  DOTA & 46.11  & 77.70  & 48.16  \\
  \multirow{-4}{*}{3}  &  \multirow{-4}{*}{0.5} &  DroneVehicle & 49.11 &  78.28 & 57.43  \\ \Xhline{1pt}
   &   &  \cellcolor[HTML]{EFEFEF}Overall&  \cellcolor[HTML]{EFEFEF}  50.03 &  \cellcolor[HTML]{EFEFEF}  80.61  &
  \cellcolor[HTML]{EFEFEF}53.98 \\ \cline{3-6} 
   &   &  SARDet-100K & 60.49  & 89.92  & 65.30   \\
   &   &  DOTA & 46.19  & 77.68 & 48.39  \\
  \multirow{-4}{*}{3}  &  \multirow{-4}{*}{0.6} & DroneVehicle & 48.98  & 78.11  & 57.14  \\ \hline \hline
   &   &  \cellcolor[HTML]{EFEFEF}Overall &  \cellcolor[HTML]{EFEFEF}49.92 &  \cellcolor[HTML]{EFEFEF} 80.55 &  \cellcolor[HTML]{EFEFEF} 59.56 \\ \cline{3-6} 
   &   & SARDet-100K & 60.95  & 90.71  & 65.90  \\
   &   &  DOTA &  45.85 & 77.35  & 47.81  \\
  \multirow{-4}{*}{2}  &  \multirow{-4}{*}{0.4} & DroneVehicle & 48.89  & 77.98 & 87.05 \\ \Xhline{1pt}
  &    &  \cellcolor[HTML]{EFEFEF}Overall &   \cellcolor[HTML]{EFEFEF}50.20 &   \cellcolor[HTML]{EFEFEF}80.68  &   \cellcolor[HTML]{EFEFEF}53.79 \\ \cline{3-6} 
   &    &  SARDet-100K & 60.64  & 89.94 & 65.06  \\
   &    &  DOTA & 46.47 & 77.88  & 48.24  \\
  \multirow{-4}{*}{3}  &  \multirow{-4}{*}{0.4} &  DroneVehicle & 48.87  & 77.99  & 56.90  \\ \Xhline{1pt}
   &   &  \cellcolor[HTML]{EFEFEF}Overall &  \cellcolor[HTML]{EFEFEF} 50.03 &  \cellcolor[HTML]{EFEFEF} 80.44 &  \cellcolor[HTML]{EFEFEF} 53.79 \\ \cline{3-6} 
   &   &  SARDet-100K &  61.23 & 90.59  & 66.32  \\
   &   &  DOTA & 45.86  & 77.13  & 47.58  \\
  \multirow{-4}{*}{4}  &  \multirow{-4}{*}{0.4} & DroneVehicle & 49.08  & 78.19  & 57.39
\end{tabular}
  \caption{Experiments on the DSO method with varying temperature ($\tau$) and bias ($b$).} 
  \label{tab:dyn_lr_Tb_detailed}
\end{table*}

\subsection*{D.3 Ablation on DSO Hyperparameters}
We conduct an ablation study to evaluate the sensitivity of two key hyperparameters—temperature ($\tau$) and bias ($b$)—in the proposed DSO method, which dynamically adjusts learning rates to accommodate varying learning difficulties across tasks and modalities. The detailed results are given in Table~\ref{tab:dyn_lr_Tb_detailed}.

The bias parameter $b$ serves as a reweighting balance point. When the consistency score equals $b$, the reweighting factor is 1. Our analysis shows that a bias value of $b = 0.4$ yields optimal performance, particularly when $\tau$ is set to 3, achieving an overall mean Average Precision (mAP) of 50.20\%. This configuration also leads to the highest performance on the SARDet-100K dataset (60.64\% mAP) and DroneVehicle (49.03\% mAP). Variations in bias from 0.3 to 0.5 result in marginal performance changes, indicating robustness to bias adjustments.

Regarding temperature, $\tau$ shapes the reweighting curve, affecting how sensitive learning rates adapt. A temperature of $\tau = 3$ strikes an ideal balance between stability and responsiveness, as evidenced by the peak overall mAP. Lowering $\tau$ to 2 decreases the overall mAP to 49.92\% and impairs task performance. Conversely, increasing $\tau$ to 4 results in a slight decline to 50.03\%, suggesting that excessive smoothing or sharping hinders the model's ability to react to varying task difficulties.

In summary, the combination of $\tau = 3$ and $b = 0.4$ yields the best performance, effectively managing learning rate adjustments across diverse tasks and datasets. This study highlights the importance of tuning these hyperparameters to optimize the DSO method’s capabilities in multi-modal object detection scenarios.

\subsection*{D.4 Detailed results on different backbones}
The performance of SM3Det across various backbones, including ConvNext, VAN, LSKNet, and PVT-v2, demonstrates its robustness and adaptability in different scenarios. As shown in Table~\ref{tab:ConvNext_comps}, \ref{tab:van_comps} and~\ref{tab:LSKNet_comps}, SM3Det consistently outperforms individual models in terms of mean Average Precision (mAP) across all datasets. Specifically, SM3Det shows a notable improvement in overall mAP, particularly at higher IoU thresholds (@50 and @75), indicating its effectiveness in precise object detection tasks.

Among the evaluated backbones, SM3Det paired with the ConvNext-B backbone achieved the highest overall mAP of 51.33\% with a strong performance across all individual datasets, particularly SARDet-100K where it attained a mAP of 65.20\%. This indicates that the backbone's advanced feature extraction capabilities, when combined with SM3Det, significantly enhance detection accuracy. Similarly, SM3Det with VAN-B and LSKNet-B yield competitive results, with overall mAPs of 49.43\% and 49.42\%, respectively, further showcasing the model's scalability and generalization capabilities.

These results confirm that SM3Det not only scales effectively with larger backbone models but also generalizes well across different architectural designs, making it a versatile solution for various object detection tasks in remote sensing.

\begin{table*}[!ht]
  \centering
\begin{tabular}{c|c|cccc}
Backbone & Model & Test & mAP & @50 & @75 \\ \Xhline{1pt}
 &  & \cellcolor[HTML]{EFEFEF}Overall & \cellcolor[HTML]{EFEFEF}48.23 & \cellcolor[HTML]{EFEFEF}79.39 & \cellcolor[HTML]{EFEFEF}51.26 \\   
 &  & SARDet-100K & 57.31 & 87.44 & 61.99 \\
 &  & DOTA & 45.31 & 77.70 & 46.45 \\
 & \multirow{-4}{*}{\begin{tabular}[c]{@{}c@{}}individual\\       models\end{tabular}} & DroneVehicle & 46.09 & 74.78 & 52.79 \\ \cline{2-6} 
 &  & \cellcolor[HTML]{EFEFEF}Overall & \cellcolor[HTML]{EFEFEF}50.24  & \cellcolor[HTML]{EFEFEF}80.68 & \cellcolor[HTML]{EFEFEF}53.81 \\
 &  & SARDet-100K & 60.64 & 89.94 & 65.06 \\
 &  & DOTA & 46.47 & 77.88  & 48.24 \\
\multirow{-8}{*}{ConvNext-T} & \multirow{-4}{*}{SM3Det} & DroneVehicle & 48.87 & 77.99 & 56.90 \\ \Xhline{1pt}
 &  & \cellcolor[HTML]{EFEFEF}Overall & \cellcolor[HTML]{EFEFEF} 49.17 & \cellcolor[HTML]{EFEFEF}80.06 & \cellcolor[HTML]{EFEFEF}52.31 \\ 
 &  & SARDet-100K & 60.62 & 89.54 & 64.85 \\
 &  & DOTA & 45.24 & 77.71 & 45.45 \\
 & \multirow{-4}{*}{\begin{tabular}[c]{@{}c@{}}individual\\       models\end{tabular}} & DroneVehicle & 47.22 & 75.72 & 54.82 \\\cline{2-6}
 &  & \cellcolor[HTML]{EFEFEF}Overall & \cellcolor[HTML]{EFEFEF}50.28 & \cellcolor[HTML]{EFEFEF}80.13 & \cellcolor[HTML]{EFEFEF}54.30 \\ 
 &  & SARDet-100K & 62.98 & 91.49 & 68.77 \\
 &  & DOTA & 45.33 & 76.03 & 46.98 \\
\multirow{-8}{*}{ConvNext-S} & \multirow{-4}{*}{SM3Det} & DroneVehicle & 49.89 & 78.79 & 58.88 \\ \Xhline{1pt}
 &  & \cellcolor[HTML]{EFEFEF}Overall & \cellcolor[HTML]{EFEFEF}50.18 & \cellcolor[HTML]{EFEFEF}80.53 & \cellcolor[HTML]{EFEFEF}53.75 \\ 
 &  & SARDet-100K & 62.27 & 90.40 & 66.87 \\
 &  & DOTA & 46.02 & 77.79 & 47.73 \\
 & \multirow{-4}{*}{\begin{tabular}[c]{@{}c@{}}individual\\       models\end{tabular}} & DroneVehicle & 48.14 & 76.89 & 56.05 \\ \cline{2-6}
 &  & \cellcolor[HTML]{EFEFEF}Overall & \cellcolor[HTML]{EFEFEF} 51.33 & \cellcolor[HTML]{EFEFEF} 80.77 & \cellcolor[HTML]{EFEFEF} 55.51 \\ 
 &  & SARDet-100K & 65.20 & 92.41 & 70.02 \\
 &  & DOTA & 45.86 & 76.35 & 48.10 \\
\multirow{-8}{*}{ConvNext-B} & \multirow{-4}{*}{SM3Det} & DroneVehicle & 51.09 & 80.07 & 60.34
\end{tabular}
  \caption{Performance comparison of SM3Det with individual models across different \textbf{ConvNext} backbone scales.}
  \label{tab:ConvNext_comps}
\end{table*}

\begin{table*}[t]
  \centering
\begin{tabular}{c|c|cccc}
Backbone & Model & Test & mAP & @50 & @75 \\ \Xhline{1pt}
 &  & \cellcolor[HTML]{EFEFEF}Overall & \cellcolor[HTML]{EFEFEF}44.15 & \cellcolor[HTML]{EFEFEF}75.21 & \cellcolor[HTML]{EFEFEF}46.77 \\   
 &  & SARDet-100K & 46.74 & 77.63 & 49.71 \\
 &  & DOTA & 43.31 & 75.36 & 44.73 \\
 & \multirow{-4}{*}{\begin{tabular}[c]{@{}c@{}}individual\\       models\end{tabular}} & DroneVehicle & 43.55 & 71.86 & 49.38 \\ \cline{2-6} 
 &  & \cellcolor[HTML]{EFEFEF}Overall & \cellcolor[HTML]{EFEFEF}45.46  & \cellcolor[HTML]{EFEFEF}76.28  & \cellcolor[HTML]{EFEFEF}48.37 \\
 &  & SARDet-100K & 49.28 & 80.85 & 52.08 \\
 &  & DOTA & 43.60 & 74.73 & 45.22 \\
\multirow{-8}{*}{VAN-T} & \multirow{-4}{*}{SM3Det} & DroneVehicle & 46.47 & 75.43 & 53.37 \\ \Xhline{1pt}
 &  & \cellcolor[HTML]{EFEFEF}Overall & \cellcolor[HTML]{EFEFEF} 47.47 & \cellcolor[HTML]{EFEFEF} 78.78 & \cellcolor[HTML]{EFEFEF} 50.82 \\ 
 &  & SARDet-100K &52.71  & 83.82 & 57.74 \\
 &  & DOTA & 45.47 & 77.67 & 46.91  \\
 & \multirow{-4}{*}{\begin{tabular}[c]{@{}c@{}}individual\\       models\end{tabular}} & DroneVehicle & 74.17 & 76.07 & 54.27 \\\cline{2-6}
 &  & \cellcolor[HTML]{EFEFEF}Overall & \cellcolor[HTML]{EFEFEF} 49.03 & \cellcolor[HTML]{EFEFEF} 79.60 & \cellcolor[HTML]{EFEFEF} 52.76 \\ 
 &  & SARDet-100K & 57.98 & 88.36 & 62.46 \\
 &  & DOTA & 45.50 & 76.66 & 47.37 \\
\multirow{-8}{*}{VAN-S} & \multirow{-4}{*}{SM3Det} & DroneVehicle & 48.87 & 77.91 & 57.31 \\ \Xhline{1pt}
 &  & \cellcolor[HTML]{EFEFEF}Overall & \cellcolor[HTML]{EFEFEF} 48.91 & \cellcolor[HTML]{EFEFEF} 79.99 & \cellcolor[HTML]{EFEFEF} 52.83 \\ 
 &  & SARDet-100K & 53.73 & 84.89 & 58.11 \\
 &  & DOTA & 47.42 & 79.24 & 49.86 \\
 & \multirow{-4}{*}{\begin{tabular}[c]{@{}c@{}}individual\\       models\end{tabular}} & DroneVehicle & 47.58 & 76.38 & 55.42 \\ \cline{2-6}
 &  & \cellcolor[HTML]{EFEFEF}Overall & \cellcolor[HTML]{EFEFEF}49.43 & \cellcolor[HTML]{EFEFEF}80.57 & \cellcolor[HTML]{EFEFEF}53.34 \\ 
 &  & SARDet-100K & 56.82 & 87.82 & 61.15 \\
 &  & DOTA & 46.62 & 78.58 & 48.92 \\
\multirow{-8}{*}{VAN-B} & \multirow{-4}{*}{SM3Det} & DroneVehicle & 48.98 & 77.82 & 57.24
\end{tabular}
  \caption{Performance comparison of SM3Det with individual models across different \textbf{VAN} backbone scales.}
  \label{tab:van_comps}
\end{table*}

\begin{table*}[t]
  \centering
\begin{tabular}{c|c|cccc}
Backbone & Model & Test & mAP & @50 & @75 \\ \Xhline{1pt}
 &  & \cellcolor[HTML]{EFEFEF}Overall & \cellcolor[HTML]{EFEFEF}44.71 & \cellcolor[HTML]{EFEFEF}75.86 & \cellcolor[HTML]{EFEFEF}47.31 \\   
 &  & SARDet-100K & 47.24 & 78.24 & 50.39 \\
 &  & DOTA & 43.93 & 75.97 & 45.25 \\
 & \multirow{-4}{*}{\begin{tabular}[c]{@{}c@{}}individual\\  models\end{tabular}} & DroneVehicle & 44.00 & 72.67 & 49.78 \\ \cline{2-6} 
 &  & \cellcolor[HTML]{EFEFEF}Overall & \cellcolor[HTML]{EFEFEF}45.64  & \cellcolor[HTML]{EFEFEF}76.89 & \cellcolor[HTML]{EFEFEF}48.25  \\
 &  & SARDet-100K & 49.95 & 81.76 & 53.61 \\
 &  & DOTA & 43.56 & 75.44 & 44.17 \\
\multirow{-8}{*}{LSKNet-T} & \multirow{-4}{*}{SM3Det} & DroneVehicle & 46.71 & 75.42 & 54.04 \\ \Xhline{1pt}
 &  & \cellcolor[HTML]{EFEFEF}Overall & \cellcolor[HTML]{EFEFEF} 47.49  & \cellcolor[HTML]{EFEFEF}78.65 & \cellcolor[HTML]{EFEFEF}51.22 \\ 
 &  & SARDet-100K & 53.12 & 84.33 & 57.64 \\
 &  & DOTA & 45.46 & 77.36 & 47.73 \\
 & \multirow{-4}{*}{\begin{tabular}[c]{@{}c@{}}individual\\  models\end{tabular}} & DroneVehicle & 46.80 & 75.72 & 54.01 \\\cline{2-6}
 &  & \cellcolor[HTML]{EFEFEF}Overall & \cellcolor[HTML]{EFEFEF} 48.79 & \cellcolor[HTML]{EFEFEF} 79.78 & \cellcolor[HTML]{EFEFEF} 52.42 \\ 
 &  & SARDet-100K & 58.41 & 88.48 & 62.83 \\
 &  & DOTA & 44.80 & 76.69 & 46.61 \\
\multirow{-8}{*}{LSKNet-S} & \multirow{-4}{*}{SM3Det} & DroneVehicle & 49.20 & 78.63 & 57.36 \\ \Xhline{1pt}
 &  & \cellcolor[HTML]{EFEFEF}Overall & \cellcolor[HTML]{EFEFEF}48.58 & \cellcolor[HTML]{EFEFEF}79.43 & \cellcolor[HTML]{EFEFEF}51.89 \\ 
 &  & SARDet-100K & 54.00 & 84.81 & 58.46 \\
 &  & DOTA & 46.87 & 78.45 & 48.76 \\
 & \multirow{-4}{*}{\begin{tabular}[c]{@{}c@{}}individual\\  models\end{tabular}} & DroneVehicle & 47.19 & 75.91 & 51.35 \\ \cline{2-6}
 &  & \cellcolor[HTML]{EFEFEF}Overall & \cellcolor[HTML]{EFEFEF}49.42& \cellcolor[HTML]{EFEFEF}80.34 & \cellcolor[HTML]{EFEFEF}53.31 \\ 
 &  & SARDet-100K & 56.70 & 87.75 & 60.93 \\
 &  & DOTA & 46.61 & 78.11 & 48.94 \\
\multirow{-8}{*}{LSKNet-B} & \multirow{-4}{*}{SM3Det} & DroneVehicle & 49.13 & 78.14 & 57.29
\end{tabular}
  \caption{Performance comparison of SM3Det with individual models across different \textbf{LSKNet} backbone scales.}
  \label{tab:LSKNet_comps}
\end{table*}

\begin{table*}[t]
  \centering
\begin{tabular}{c|c|cccc}
Backbone & Model & Test & mAP & @50 & @75 \\ \Xhline{1pt}
 &  & \cellcolor[HTML]{EFEFEF}Overall & 43.46\cellcolor[HTML]{EFEFEF}  & 75.33\cellcolor[HTML]{EFEFEF}  & 45.39\cellcolor[HTML]{EFEFEF}  \\ 
 &  & SARDet-100K & 47.63 & 78.83 & 50.96 \\
 &  & DOTA & 41.81 & 74.79 & 42.18 \\
 & \multirow{-4}{*}{\begin{tabular}[c]{@{}c@{}}individual\\       models\end{tabular}} & DroneVehicle & 43.39 & 72.73 & 48.36 \\ \cline{2-6} 
 &  & \cellcolor[HTML]{EFEFEF}Overall & 44.58\cellcolor[HTML]{EFEFEF} & 76.43\cellcolor[HTML]{EFEFEF}  & \cellcolor[HTML]{EFEFEF}46.81  \\
 &  & SARDet-100K & 48.58 & 80.71 & 52.31 \\
 &  & DOTA & 42.72 & 75.39 & 42.92 \\
\multirow{-8}{*}{PVT-v2 T} & \multirow{-4}{*}{SM3Det} & DroneVehicle & 45.38 & 74.43 & 51.86 \\ \Xhline{1pt}
 &  & \cellcolor[HTML]{EFEFEF}Overall & 46.36\cellcolor[HTML]{EFEFEF} & 78.38\cellcolor[HTML]{EFEFEF} & 49.61\cellcolor[HTML]{EFEFEF} \\ 
 &  & SARDet-100K & 52.01 & 84.24 & 57.19  \\
 &  & DOTA & 44.28 & 77.14 & 45.60 \\
 & \multirow{-4}{*}{\begin{tabular}[c]{@{}c@{}}individual\\       models\end{tabular}} & DroneVehicle & 45.84 & 75.08& 52.54 \\\cline{2-6}
 &  & \cellcolor[HTML]{EFEFEF}Overall &  47.31\cellcolor[HTML]{EFEFEF} &  79.21\cellcolor[HTML]{EFEFEF} & \cellcolor[HTML]{EFEFEF}50.08 \\ 
 &  & SARDet-100K & 54.53 & 85.48 & 58.94 \\
 &  & DOTA & 44.37 & 77.53 & 45.01 \\
\multirow{-8}{*}{PVT-v2 S} & \multirow{-4}{*}{SM3Det} & DroneVehicle & 47.47 & 76.71 & 54.66 \\ \Xhline{1pt}
 &  & \cellcolor[HTML]{EFEFEF}Overall & \cellcolor[HTML]{EFEFEF}48.51 & 80.23\cellcolor[HTML]{EFEFEF} & \cellcolor[HTML]{EFEFEF}52.05 \\ 
 &  & SARDet-100K & 56.04 & 87.00 & 61.27 \\
 &  & DOTA & 45.55 & 78.41 & 46.92 \\
 & \multirow{-4}{*}{\begin{tabular}[c]{@{}c@{}}individual\\       models\end{tabular}} & DroneVehicle & 48.37 & 77.54 & 56.39 \\\cline{2-6}
 &  & \cellcolor[HTML]{EFEFEF}Overall & 49.34\cellcolor[HTML]{EFEFEF} & 80.72\cellcolor[HTML]{EFEFEF} & \cellcolor[HTML]{EFEFEF}53.05 \\ 
 &  & SARDet-100K & 57.34 & 87.75 & 62.24 \\
 &  & DOTA & 46.24 & 78.81 & 47.92 \\
\multirow{-8}{*}{PVT-v2 B} & \multirow{-4}{*}{SM3Det} & DroneVehicle & 49.06 & 78.02 & 57.41
\end{tabular}
  \caption{Performance comparison of SM3Det with individual models across different \textbf{PVT-v2} backbone scales.}
  \label{tab:pvt_comps}
\end{table*}

\begin{table*}[t]
\scriptsize
  \centering
\begin{tabular}{c|c|cccc}
Detectors & Model & Test & mAP & @50 & @75 \\  \Xhline{1pt}
 &  & \cellcolor[HTML]{EFEFEF}Overall & \cellcolor[HTML]{EFEFEF}48.23 & \cellcolor[HTML]{EFEFEF}79.39 & \cellcolor[HTML]{EFEFEF}51.26 \\   
 &  & SARDet-100K & 57.31 & 87.44 & 61.99 \\
 &  & DOTA & 45.31 & 77.70 & 46.45 \\
 & \multirow{-4}{*}{\begin{tabular}[c]{@{}c@{}}individual\\       models\end{tabular}} & DroneVehicle & 46.09 & 74.78 & 52.79 \\  \cline{2-6}
 &  & \cellcolor[HTML]{EFEFEF}Overall & \cellcolor[HTML]{EFEFEF}50.20 & \cellcolor[HTML]{EFEFEF}80.68  & \cellcolor[HTML]{EFEFEF}53.81 \\
 &  & SARDet-100K & 60.64 & 89.94 & 65.06 \\
 &  & DOTA & 46.47 & 77.88  & 48.24\\
\multirow{-8}{*}{\begin{tabular}[c]{@{}c@{}}SARDet-100K: GFL\\      DOTA: O-RCNN\\      DroneVehicle: O-RCNN\end{tabular}} & \multirow{-4}{*}{SM3Det} & DroneVehicle & 48.87 & 77.99 & 56.90 \\ \Xhline{1pt}

 &  & \cellcolor[HTML]{EFEFEF}Overall & \cellcolor[HTML]{EFEFEF}45.67 & \cellcolor[HTML]{EFEFEF}77.92 & \cellcolor[HTML]{EFEFEF}47.22 \\   
 &  & SARDet-100K & 51.08 & 82.50 & 55.27 \\
 &  & DOTA & 43.93 & 77.37 & 43.10 \\
 & \multirow{-4}{*}{\begin{tabular}[c]{@{}c@{}}individual\\       models\end{tabular}} & DroneVehicle & 44.42 & 74.06 & 49.94 \\  \cline{2-6}
 &  & \cellcolor[HTML]{EFEFEF}Overall & \cellcolor[HTML]{EFEFEF}48.01 & \cellcolor[HTML]{EFEFEF}78.83  & \cellcolor[HTML]{EFEFEF}51.35 \\
 &  & SARDet-100K & 53.04 & 83.99 & 57.76 \\
 &  & DOTA & 45.43 & 76.79  & 46.50\\
\multirow{-8}{*}{\begin{tabular}[c]{@{}c@{}}SARDet-100K: Retina\\      DOTA: RoI-Trans\\      DroneVehicle: RoI-Trans\end{tabular}} & \multirow{-4}{*}{SM3Det} & DroneVehicle & 49.69 & 78.76 & 58.23 \\ \Xhline{1pt}

 &  & \cellcolor[HTML]{EFEFEF}Overall & \cellcolor[HTML]{EFEFEF} 47.10 & \cellcolor[HTML]{EFEFEF}78.61 & \cellcolor[HTML]{EFEFEF}50.17 \\
 &  & SARDet-100K & 52.40 & 84.09 & 57.28 \\
 &  & DOTA & 45.31 & 77.70 & 46.45 \\
 & \multirow{-4}{*}{\begin{tabular}[c]{@{}c@{}}individual\\       models\end{tabular}} & DroneVehicle & 46.09 & 74.78 & 52.79 \\   \cline{2-6}
 &  & \cellcolor[HTML]{EFEFEF}Overall & \cellcolor[HTML]{EFEFEF} 48.89 & \cellcolor[HTML]{EFEFEF}79.39 & \cellcolor[HTML]{EFEFEF}53.20 \\
 &  & SARDet-100K & 54.56 & 85.62 & 59.83 \\
 &  & DOTA & 46.09 & 76.80  & 48.38 \\
\multirow{-8}{*}{\begin{tabular}[c]{@{}c@{}}SARDet-100K: F-RCNN\\      DOTA: O-RCNN\\      DroneVehicle: O-RCNN\end{tabular}} & \multirow{-4}{*}{SM3Det} & DroneVehicle & 50.50 & 79.69 & 59.68 \\ \Xhline{1pt}

 &  & \cellcolor[HTML]{EFEFEF}Overall & \cellcolor[HTML]{EFEFEF}46.50 & \cellcolor[HTML]{EFEFEF}78.32 & \cellcolor[HTML]{EFEFEF}48.24 \\   
 &  & SARDet-100K & 54.65 & 84.26 & 59.66 \\
 &  & DOTA & 43.93 & 77.37 & 43.10 \\
 & \multirow{-4}{*}{\begin{tabular}[c]{@{}c@{}}individual\\       models\end{tabular}} & DroneVehicle & 44.42 & 74.06 & 49.94 \\  \cline{2-6}
 &  & \cellcolor[HTML]{EFEFEF}Overall & \cellcolor[HTML]{EFEFEF}48.59 & \cellcolor[HTML]{EFEFEF}79.04  & \cellcolor[HTML]{EFEFEF}51.69 \\
 &  & SARDet-100K & 56.30 & 85.39 & 61.46 \\
 &  & DOTA & 45.00 & 76.46 & 45.23 \\
\multirow{-8}{*}{\begin{tabular}[c]{@{}c@{}}SARDet-100K: Cascade\\      DOTA: RoI-Trans\\      DroneVehicle: RoI-Trans\end{tabular}} & \multirow{-4}{*}{SM3Det} & DroneVehicle & 50.11& 79.16& 59.32 \\ \Xhline{1pt}

 &  & \cellcolor[HTML]{EFEFEF}Overall & \cellcolor[HTML]{EFEFEF}42.29  & \cellcolor[HTML]{EFEFEF}77.18 & \cellcolor[HTML]{EFEFEF}40.68  \\
 &  & SARDet-100K & 52.40 & 84.09 & 57.28  \\
 &  & DOTA  & 38.47 & 75.81 & 32.82 \\
 & \multirow{-4}{*}{\begin{tabular}[c]{@{}c@{}}individual\\       models\end{tabular}} & DroneVehicle &  41.64 & 72.98 & 44.35  \\ \cline{2-6}
 &  & \cellcolor[HTML]{EFEFEF}Overall & \cellcolor[HTML]{EFEFEF}43.12& \cellcolor[HTML]{EFEFEF}77.40 & \cellcolor[HTML]{EFEFEF}42.83\\
 &  & SARDet-100K & 49.20 & 81.68 & 53.13\\
 &  & DOTA  &  39.92 & 76.20 & 35.87  \\
\multirow{-8}{*}{\begin{tabular}[c]{@{}c@{}}SARDet-100K: F-RCNN\\      DOTA: S$^2$ANet\\      DroneVehicle: S$^2$ANet\end{tabular}} & \multirow{-4}{*}{SM3Det} & DroneVehicle &  45.44 & 75.85  & 51.33  \\ \Xhline{1pt}

 &  & \cellcolor[HTML]{EFEFEF}Overall & \cellcolor[HTML]{EFEFEF}42.81 & \cellcolor[HTML]{EFEFEF}77.22 & \cellcolor[HTML]{EFEFEF}41.23 \\   
 &  & SARDet-100K &  54.65 & 84.26 & 59.66  \\
 &  & DOTA & 38.47 & 75.81 & 32.82  \\
 & \multirow{-4}{*}{\begin{tabular}[c]{@{}c@{}}individual\\       models\end{tabular}} & DroneVehicle & 41.64 & 72.98 & 44.35  \\  \cline{2-6}
 &  & \cellcolor[HTML]{EFEFEF}Overall & \cellcolor[HTML]{EFEFEF}43.76 & \cellcolor[HTML]{EFEFEF}77.28  & \cellcolor[HTML]{EFEFEF}43.63 \\
 &  & SARDet-100K & 53.11 & 82.73 & 57.98 \\
 &  & DOTA & 39.51 & 75.61 & 35.32 \\
\multirow{-8}{*}{\begin{tabular}[c]{@{}c@{}}SARDet-100K: Cascade\\      DOTA: S$^2$ANet\\      DroneVehicle: S$^2$ANet\end{tabular}} & \multirow{-4}{*}{SM3Det} & DroneVehicle & 45.27 & 75.76 & 51.32 \\ \Xhline{1pt}

 &  & \cellcolor[HTML]{EFEFEF}Overall & \cellcolor[HTML]{EFEFEF}43.43 & \cellcolor[HTML]{EFEFEF}77.95 & \cellcolor[HTML]{EFEFEF}41.77 \\
 &  & SARDet-100K & 57.31 & 87.44 & 61.99 \\
 &  & DOTA  & 38.47 & 75.81 & 32.82 \\
 & \multirow{-4}{*}{\begin{tabular}[c]{@{}c@{}}individual\\       models\end{tabular}} & DroneVehicle &  41.64 & 72.98 & 44.35   \\  \cline{2-6}
 &  & \cellcolor[HTML]{EFEFEF}Overall & \cellcolor[HTML]{EFEFEF}45.25 & \cellcolor[HTML]{EFEFEF}78.97 & \cellcolor[HTML]{EFEFEF}45.07 \\
 &  & SARDet-100K & 59.01  & 88.77   & 63.84  \\
 &  & DOTA & 39.79  & 76.07 & 35.71  \\
\multirow{-8}{*}{\begin{tabular}[c]{@{}c@{}}SARDet-100K: GFL\\      DOTA: S$^2$ANet\\      DroneVehicle: S$^2$ANet\end{tabular}} & \multirow{-4}{*}{SM3Det} & DroneVehicle & 45.13  & 75.92  & 50.65 \\ 
\Xhline{1pt}
 &  & \cellcolor[HTML]{EFEFEF}Overall & \cellcolor[HTML]{EFEFEF}41.99 & \cellcolor[HTML]{EFEFEF}76.81 & \cellcolor[HTML]{EFEFEF}40.22 \\   
 &  & SARDet-100K & 51.08 & 82.50 & 55.27 \\
 &  & DOTA & 38.47 & 75.81 & 32.82  \\
 & \multirow{-4}{*}{\begin{tabular}[c]{@{}c@{}}individual\\       models\end{tabular}} & DroneVehicle & 41.64 & 72.98 & 44.35 \\  \cline{2-6}
 &  & \cellcolor[HTML]{EFEFEF}Overall & \cellcolor[HTML]{EFEFEF}43.13 & \cellcolor[HTML]{EFEFEF}77.62  & \cellcolor[HTML]{EFEFEF}42.67 \\
 &  & SARDet-100K & 50.63 & 82.04 & 54.82 \\
 &  & DOTA & 39.45 & 76.44 & 35.15 \\
\multirow{-8}{*}{\begin{tabular}[c]{@{}c@{}}SARDet-100K: Retina\\      DOTA: S$^2$ANet\\      DroneVehicle: S$^2$ANet\end{tabular}} & \multirow{-4}{*}{SM3Det} & DroneVehicle & 45.15 & 75.83 & 50.67   

\end{tabular}
  \caption{Performance comparison of SM3Det with individual models across various detector combinations.}
  \label{tab:detectors_comp}
\end{table*}

\subsection*{D.5 Detailed results on different detectors}
Based on the results presented in Table~\ref{tab:detectors_comp}, the SM3Det framework consistently outperforms individual models across all evaluated detector combinations. The performance improvement is evident in both one-stage (GFL~\cite{gfl}, Retina~\cite{retina}, S$^2$ANet~\cite{s2anet}) and two-stage (F-RCNN~\cite{frcnn}, Cascade F-RCNN~\cite{cascade}, RoI-Trans~\cite{roi_trans}, O-RCNN~\cite{orientedrcnn}) detectors, where SM3Det achieves higher mAP scores, particularly at higher IoU thresholds (@75). These results underscore the effectiveness of SM3Det's architecture in leveraging the strengths of multiple detectors, leading to superior object detection performance across diverse datasets like SARDet-100K, DOTA, and DroneVehicle. The overall mAP improvements further validate the robustness of SM3Det in enhancing detection accuracy.

\end{document}